\RequirePackage{fix-cm}
\documentclass[twocolumn]{svjour3}          
\smartqed  
\usepackage{graphicx}
\usepackage{amssymb}
\usepackage[cmex10]{amsmath}
\usepackage{array}
\usepackage{url}
\usepackage{color}
\usepackage[tight,footnotesize]{subfigure}
\usepackage{natbib}
\usepackage{hyphenat}
\usepackage[pdftex,
CJKbookmarks=true, bookmarksnumbered=true, bookmarksopen=true,
colorlinks=true,
pdfborder=001,
citecolor=blue, linkcolor=blue, anchorcolor=red, urlcolor=black
]{hyperref}
\usepackage{bbm,dsfont}
\usepackage{indentfirst} 
\usepackage{booktabs}

\usepackage[dvipsnames]{xcolor} 
\usepackage{caption}
\usepackage{subcaption}
\usepackage{multirow}
\usepackage{tabularx}
\usepackage{adjustbox}
\usepackage{pifont}
\usepackage[ruled,vlined]{algorithm2e}
\usepackage[skip=2pt]{caption}
\usepackage{marvosym}

\journalname{International Journal of Computer Vision}
\definecolor{darkmag}{rgb}{0.55,0,0.55}
\begin{document}
\begin{sloppypar}

\title{Exploring Homogeneous and Heterogeneous Consistent Label Associations for Unsupervised Visible-Infrared Person ReID}

\titlerunning{Short form of title}        


\author{Lingfeng He \and
        De Cheng\textsuperscript{\Letter} \and
        Nannan Wang\textsuperscript{\Letter} \and
        Xinbo Gao \and
}

\authorrunning{Short form of author list} 

\institute{
Lingfeng He \at
Xidian University, Xi'an 710071, China \\
\email{\href{lfhe@stu.xidian.edu.cn}{lfhe@stu.xidian.edu.cn}} 
\and
De Cheng\textsuperscript{\Letter}, Corresponding authors \at
Xidian University, Xi'an 710071, China \\
\email{\href{dcheng@xidian.edu.cn}{dcheng@xidian.edu.cn}}
\and
Nannan Wang \at
Xidian University, Xi'an 710071, China \\
\email{\href{nnwang@xidian.edu.cn}{nnwang@xidian.edu.cn}}
\and
Xinbo Gao \at
Chongqing University of Posts and Telecommunications, Chongqing 400065, China \\
\email{\href{gaoxb@cqupt.edu.cn}{gaoxb@cqupt.edu.cn}}
}

\date{Received: date / Accepted: date}

\maketitle

\begin{abstract}

Unsupervised visible-infrared person re-identification (USL-VI-ReID) endeavors to retrieve pedestrian images of the same identity from different modalities without annotations. 
While prior work focuses on establishing cross-modality pseudo-label associations to bridge the modality-gap, they ignore maintaining the instance-level homogeneous and heterogeneous consistency between the feature space and the pseudo-label space, resulting in coarse associations.
In response, we introduce a Modality-Unified Label Transfer (MULT) module that simultaneously accounts for both homogeneous and heterogeneous fine-grained instance-level structures, yielding high-quality cross-modality label associations. 
It models both homogeneous and heterogeneous affinities, leveraging them to
quantify the inconsistency between the pseudo-label space and the feature space,
subsequently minimizing it.
The proposed MULT ensures that the generated pseudo-labels maintain alignment across modalities while upholding structural consistency within intra-modality.
Additionally, a straightforward plug-and-play Online Cross-memory Label Refinement (OCLR) module is proposed to further mitigate the side effects of noisy pseudo-labels while simultaneously aligning different modalities, coupled with an Alternative Modality-Invariant Representation Learning (AMIRL) framework.
Experiments demonstrate that our proposed method outperforms existing state-of-the-art USL-VI-ReID methods, highlighting the superiority of our MULT in comparison to other cross-modality association methods.
Code is available at \href{https://github.com/FranklinLingfeng/code_for_MULT}{https://github.com/FranklinLingfeng/code\_for\_MULT}.

\end{abstract}

\section{Introduction}

Visible-infrared person re-identification (VI-ReID) \cite{VI-ReID1, VI-ReID2, SYSU-MM01, agw, DDAG, PGM} aims at retrieving the same person from a set of visible/infrared gallery images when given an image from another modality. It has garnered growing interest due to its practical applications in intelligent surveillance systems. Existing VI-ReID methods have achieved remarkable performance with deep neural networks \cite{DNN1, DNN2, DNN3, AlignGAN, CIFT, MAUM}. However, these works are based on datasets with modality-shared annotations, which are labor-intensive and time-consuming to obtain in real-world scenarios. To relieve such issues, we investigate the unsupervised solution for VI-ReID.

\begin{figure}
\centering
\includegraphics[width=0.48\textwidth]{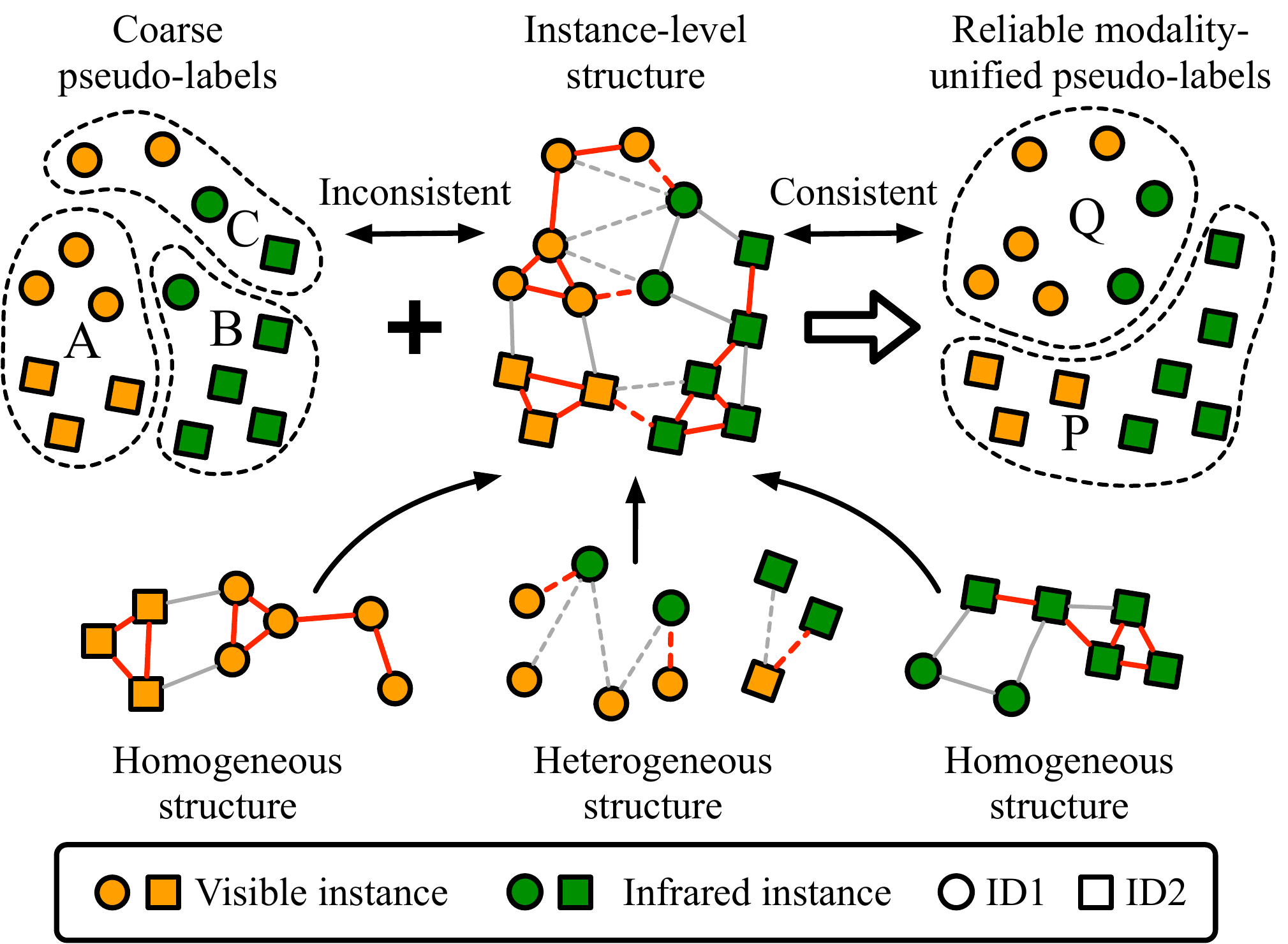}
\caption{
Illustration of our idea. 
Different colors denote different modalities, and different shapes denote different identities. 
\textcolor{red}{The red lines} represent \textbf{higher affinities} and \textcolor{gray}{the gray lines} represent \textbf{lower affinities}. 
Our Modality-Unified Label Transfer takes into account instance-level structures to establish homogeneous and heterogeneous structurally consistent label associations 
and generate reliable modality-unified pseudo-labels for network training.
}\label{motivation}
\vspace{-6mm}
\end{figure}

Although there are several unsupervised single-modality ReID methods that have achieved excellent performance ($e.g.$, Cluster-Contrast \cite{ClusterContrast}, ISE \cite{ISE}, PPLR \cite{PPLR}), the direct application of these methods to VI-ReID scenarios presents a formidable challenge due to the substantial modality-gap. 
Images of the same pedestrian from different modalities cannot be assigned the same pseudo-label using conventional clustering-based methods. 
Therefore, associating cross-modality pseudo-labels is necessary for unsupervised VI-ReID. Several attempts \cite{OTLA, DOTLA, MBCCM, PGM} have been made to associate the cross-modality pseudo-labels. PGM \cite{PGM} and MBCCM \cite{MBCCM} adopt graph matching to interlink cross-modality clusters from the global perspective.
However, they overlook the complicated, fine-grained structural information at the instance level, consequently resulting in coarse associations. 
To utilize the instance-level information, OTLA \cite{OTLA} formulates the label assignment between instance and cross-modality clusters as an Optimal Transport (OT) problem. 
Nevertheless, it neglects the homogeneous structural consistency, leading to a large amount of intra-cluster instances in one modality being dispersed across multiple clusters in another modality. 
Based on the above analysis, we utilize instance-level pairwise relationships to establish reliable cross-modality label associations that maintain both homogeneous and heterogeneous structural consistency (as shown in Fig.\ref{motivation}).

Specifically, we propose a Modality-Unified Label Transfer (MULT) module (Fig.\ref{framework} (a)), which exploits the full potential of both homogeneous and heterogeneous instance-level structures to associate cross-modality pseudo-labels.
To begin, our MULT excavates homogeneous and heterogeneous structural information by modeling affinities derived from pairwise instance relationships in feature space.
These affinities are then utilized to define both homogeneous and heterogeneous inconsistency between the pseudo-label space and the feature space from a global perspective. 
Subsequently, MULT transfers the pseudo-labels guided by the calculated affinities, with the primary aim of minimizing the inconsistency terms. 
During the label transfer process, each instance communicates its pseudo-label information with both its intra-modality and cross-modality counterparts. 
Such transfer strategy leverages detailed instance-level relationships, facilitating more precise associations compared to the direct associations of clusters.
Simultaneously, the pseudo-labels preserve the homogeneous structure in feature space by minimizing the homogeneous inconsistency terms.
Furthermore, the transferred soft pseudo-labels involve information between instances and multiple intra-modality and cross-modality clusters and thus are more suitable for mining discriminative features compared to hard labels.
Extensive experiments indicate that our MULT provides more suitable supervision signals for training compared to other cross-association methods.

To achieve modality alignment based on pseudo-labels derived from MULT, we introduce an Online Cross-memory Label Refinement (OCLR) module (Fig.\ref{framework} (c)), complemented by an Alternative Modality-Invariant Representation Learning (AMIRL) framework (Fig.\ref{framework} (b)).
Specifically, our OCLR module is a straightforward yet effective plug-and-play component designed to alleviate the impact of the inevitable noisy pseudo-labels while further reducing the modality-gap. Specifically,
it learns self-consistency among the predictions from multi-memory prototypes.
{
Our AMIRL framework conducts contrastive learning based on both intra-modality and cross-modality memory banks.
Two auxiliary memory banks are constructed to collaboratively learn the intra-modality structure with the initial two intra-modality memory banks, which play a role in mutually correcting each other. 
Furthermore, an alternative training scheme is proposed to multigate the influence of the inconsistency between the pseudo-labels from MULT in different directions.}
Experimental results highlight the effectiveness of our OCLR, showcasing its applicability in various cross-modality label association methods.
Our main contributions can be summarized as follows:

\begin{itemize} 

\item We propose a MULT module that considers instance-level context structures to provide homogeneous and heterogeneous consistent cross-modality pseudo-label associations for network training. The generated pseudo-labels exhibit cross-modality alignment while containing rich intra-modality information.

\item We design a straightforward plug-and-play OCLR module for learning cross-memory self-consistency online, coupled with an AMIRL framework that fully exploits the supervision signals from MULT. The OCLR module effectively alleviates the side-effects of the noisy labels while mitigating the modality-gap.


\item Extensive experimental results demonstrate the effectiveness of the proposed method, showing higher-quality label associations across-modality instances, and superior recognition performances to state-of-the-art USL-VI-ReID methods on SYSU-MM01 and RegDB datasets.


\end{itemize}

\section{Related Work}

\subsection{Supervised VI-ReID}

Supervised VI-ReID pays attention to retrieving pedestrian images from both visible and infrared cameras. Existing methods mainly focus on aligning two modalities at the image-level and feature-level. Some methods \cite{AlignGAN, cmGAN, hi-cmd} bridge the modality-gap by transferring images from one modality to the other based on Generative Adversarial Networks (GANs). However, these methods face the issue of high computational cost and the inevitable noise in generated images, thus degenerating the model performance. CA \cite{CA} proposes a simple random channel argumentation for visible images to bridge the gap between cross-modality images. For aligning modalities at feature-level, some methods design novel network structures based on two-stream CNN with deep metric learning \cite{MAUM, MPANet, DDAG, DEEN, CAL} to excavate modality-invariant features. DDAG \cite{DDAG} mines both intra-modality part-level and cross-modality graph-level contextual information through a dynamic dual-attention aggregation learning method.
MPANet \cite{MPANet} proposes a modality alleviation module and a pattern alignment module to extract part-level discriminative features.
PartMix \cite{PartMix} synthesizes positive and negative by mixing the part descriptors across modalities to enhance part-level feature learning. However, the aforementioned methods rely heavily on human-annotated cross-modality associations, which require expensive labor costs and are not always available in real scenarios. Thus we focus on learning modality-invariant features without human annotations.

\subsection{Unsupervised Single-Modality ReID}

Unsupervised single-modality ReID aims to learn discriminative representations for unannotated person images. Recent unsupervised methods \cite{USL1, USL2, USL3, USL4, USL5, USL6, USL7, ClusterContrast, ISE} follow a self-training pipeline that alternates between generating pseudo-labels and training the network. 
The mainstream methods for unsupervised ReID include pseudo-label refinement-based methods and memory bank-based methods.
To alleviate the impact of noisy labels, MMT \cite{MMT} utilizes the Mean-Teacher \cite{mean-teacher} model to revise pseudo-labels online.
RLCC \cite{RLCC} propagates pseudo-labels between different iterations through cluster consensus and generates temporally consistent pseudo-labels.
PPLR \cite{PPLR} proposes refining the pseudo-labels of global features by ensembling the predictions of part features based on the global-part agreement score.
Memory bank-based methods establish memories capable of managing multi-prototypes and facilitate contrastive learning during training.
SPCL \cite{SPCL} constructs a hybrid memory that maintains both instance-level and cluster-level prototypes with a self-paced learning strategy. 
Cluster-Contrast \cite{ClusterContrast} and ISE \cite{ISE} store a unique prototype for each cluster to preserve the updating consistency.
DCMIP \cite{DCMIP} manages discrepant cluster memories and a multi-instance memory to excavate multifaceted information within clusters.

Following the memory-based methods, we construct intra-modality and cross-modality memory banks to perform contrastive learning for heterogeneous features.

\subsection{Unsupervised VI-ReID}

Unsupervised VI-ReID aims to learn modality-invariant and identity-discriminative features for cross-modality images without annotations. Existing methods \cite{H2H, ADCA, MBCCM, DOTLA, PGM} mainly focus on associating cross-modality pseudo-labels. 
H2H \cite{H2H} proposes an ISML loss aimed at enhancing alignment between reliable cross-modality instances.
ADCA \cite{ADCA} identifies highly associated cross-modality features and aggregates their corresponding memory prototypes according to instance-level pairwise similarities.
To avoid biased label associations, 
OTLA \cite{OTLA} establishes an optimal transport problem to assign cross-modality pseudo-labels uniformly for instances. 
To model the cluster-level cross-modality relationships from a global perspective, PGM \cite{PGM} and MBCCM \cite{MBCCM} construct weighted bipartite graphs to associate cross-modality clusters and assign shared pseudo-labels for heterogeneous instances.
CCLNet \cite{CCLNet} leverages the powerful semantic information from CLIP to provide richer supervision signals by incorporating a prompt learning stage.
GRU \cite{GRU} proposes a CAE module to embed the information of hierarchical domain memories while achieving remarkable performance.

In contrast to the above-mentioned methods, our MULT simultaneously utilizes the homogeneous and heterogeneous instance-level structures to provide reliable cross-modality label associations.

\begin{figure*}
\centering
\includegraphics[width=1.0\textwidth]{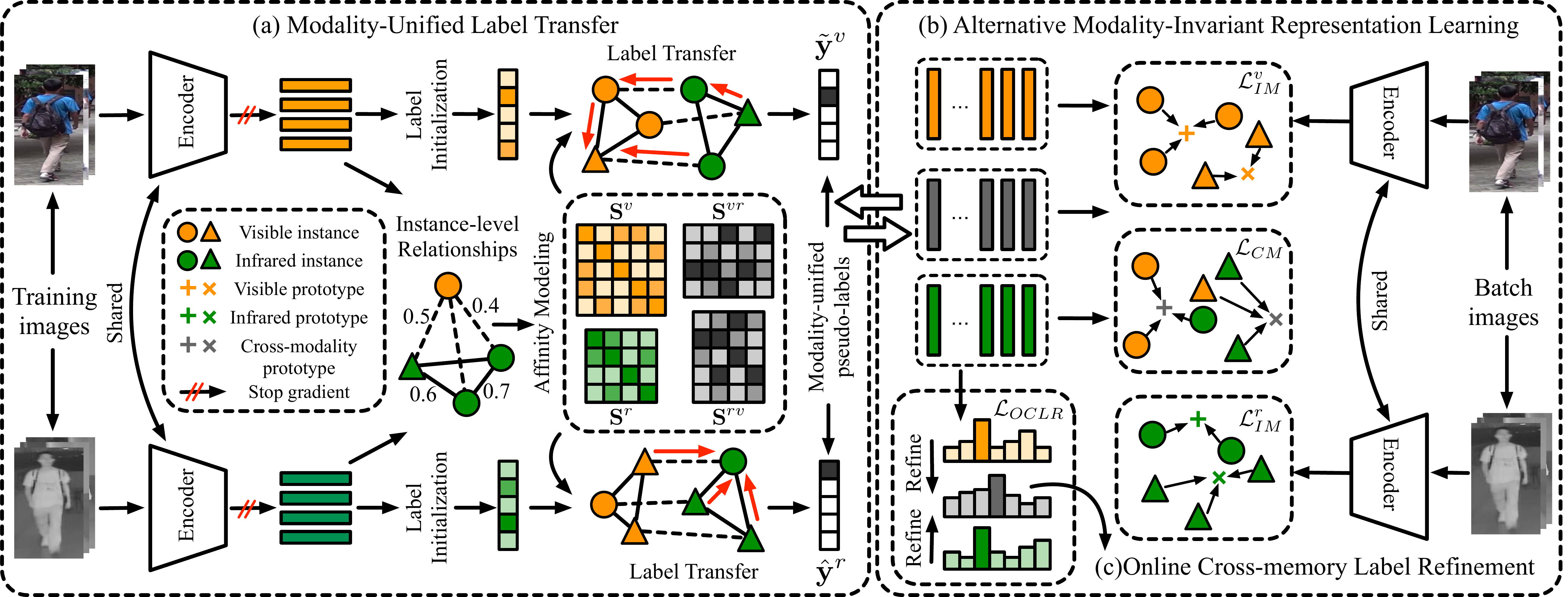}\\
\caption{Framework of our proposed method. Different colors indicate different modalities. Our method alternates pseudo-label generation
(Modality-Unified Label Transfer (MULT (a), described in Sec.\ref{MULT})) and 
network training (including Alternative Modality-Invariant Representation Learning (AMIRL (b), described in Sec.\ref{MIRL}) and Online Cross-memory Label Refinement (OCLR (c), described in Sec.\ref{OCLR})).
MULT provides homogeneous and heterogeneous consistent pseudo-labels as supervision signals. 
During training, AMIRL leverages memory banks to perform contrastive learning with an alternative scheme
and OCLR utilizes predictions from different memories to alleviate the effect of the noisy labels.
}\label{framework}
\vspace{-3mm}
\end{figure*}

\subsection{Affinity-Based Person ReID}

In the person ReID task, some works \cite{graph1, graph2, graph3, SSFT, cm-SSFT, CIFT, DDAG} pay attention to the detailed relationships between pairwise instances. 
SSFT \cite{SSFT} proposes a feature transfer module to facilitate the optimization of group-wise similarities for single-modality ReID. 
cm-SSFT \cite{cm-SSFT} models both intra-modality and cross-modality affinities to generate modality-shared features from modality-specific features. 
CIFT \cite{CIFT} proposes a novel graph module to simulate the unbalanced modality distribution. 

Inspired by the widely-used GCN \cite{GCN} and the above affinity-based methods, we model affinities in feature space to fully exploit instance-level relationships. Then we aim to obtain pseudo-labels that preserve instance-level contextual information.

\section{Methodology}
\label{sec:method}


Given a visible-infrared dataset $\mathcal{X} = \left\{ {\mathcal{X}}^v, {\mathcal{X}}^r \right\}$, ${\mathcal{X}}^v = \left\{\mathbf{x}^v_i | i=1, 2, \cdots, N^v\right\}$ and
${\mathcal{X}}^r = \left\{\mathbf{x}^r_i | i=1, 2, \cdots, N^r\right\}$ represents the visible and infrared datasets with $N^v$ and $N^r$ images, respectively. In the context of the USL-VI-ReID task, our objective is to train a deep neural network $f_{\theta}(\cdot)$ to project an image $\mathbf{x}_i$ from the dataset $\mathcal{X}$ into an embedding space $\mathcal{F}_{\theta}$ and derive a d-dimensional modality-invariant representation $\mathbf{f}_i = f_{\theta}(\mathbf{x}_i) \in \mathbb{R}^{d}$. 


We employ the two stream encoder $f_{\theta}(\cdot)$ ($e.g.$, ResNet50 \cite{resnet}) as backbone to extract visible features $\left\{\mathbf{f}^v_i | i=1, 2, \cdots, N^v\right\}$ and infrared features $\left\{\mathbf{f}^r_i | i=1, 2, \cdots, N^r\right\}$.
In the initial training stage, 
following \cite{ADCA, MBCCM, PGM}, 
we employ the Dual-Contrastive Learning (DCL) framework \cite{ADCA} to facilitate intra-modality contrastive learning as our baseline.


Our proposed method is employed during the second training stage.
The framework is illustrated in Fig.\ref{framework}. 
Following the well-developed unsupervised methods \cite{SPCL, ClusterContrast, ISE, ADCA}, we alternate between pseudo-label generation and network training.
During the pseudo-label generation stage, following \cite{ADCA, PGM, MBCCM}, we first utilize DBSCAN \cite{1996DBSCAN} to cluster features. Two intra-modality memory banks $\tilde{\mathbf{M}}^v \in \mathbb{R}^{K^v \times d}$ and $\tilde{\mathbf{M}}^r \in \mathbb{R}^{K^r \times d}$ are initialized by the cluster centroids of their corresponding modalities, where $K^v$ and $K^r$ denotes the number of clusters in the visible modality and the infrared modality, respectively.
Then the proposed Modality-Unified Label Transfer (MULT, Fig.\ref{framework} (a)) establishes reliable cross-modality label associations and generates soft modality-unified pseudo-labels for network training.
During the training stage, we propose an Alternative Modality-Invariant Representation learning (AMIRL, Fig.\ref{framework} (b)) framework to fully leverage the supervisory signals from MULT through contrastive learning, coupled with an alternative scheme.
Additionally, an Online Cross-memory Label refinement (OCLR, Fig.\ref{framework} (c)) module is proposed to further alleviate the influence of noisy labels while reducing the modality discrepancy. 
The entire framework includes two training modes, $i.e.$, V-based and R-based, where V-based denotes we utilize pseudo-labels in the visible label space to guide network training, and vice versa. 

\subsection{Modality-Unified Label Transfer}\label{MULT}

Motivated by the affinity-based ReID methods \cite{SSFT, cm-SSFT, CIFT}, Our MULT models instance-wise affinities to generate structurally consistent pseudo-labels.
In MULT, every instance holds two types of pseudo-labels, $i.e.$, the intra-modality label (labels from the label space corresponding to the instance's modality) and the cross-modality label (labels from the label space corresponding to another modality).
Formally, 
We use $\mathbf{y}^e = \{\tilde{\mathbf{y}}^e, \hat{\mathbf{y}}^e\}$ to denote the pseudo-labels, where $\tilde{\mathbf{y}}^e$ denotes the intra-modality labels and $\hat{\mathbf{y}}^e$ denotes the cross-modality labels. $e = \{v, r\}$ indicates visible and infrared modality, respectively, and $\mathbf{y}^e_i$ denotes the pseudo-label of $i$-th instance in modality $e$.
($e.g.$, $\tilde{\mathbf{y}}^v$ denotes pseudo-labels for visible instances in visible label space and $\hat{\mathbf{y}}^v$ denotes pseudo-labels for visible instances in infrared label space).

Our MULT includes two directions, $i.e.$, V2R and R2V. 
The V2R MULT involves the transfer between visible intra-modality labels 
$\tilde{\mathbf{y}}^v$ and infrared cross-modality labels $\hat{\mathbf{y}}^r$ within the visible label space, and verse vice.
For convenience, we only describe the V2R MULT in detail.







\noindent\textbf{Affinity Modeling.} 
To incorporate instance-level relationships, we model the homogeneous and heterogeneous affinities, denoted as $\mathbf{S}^{ho(e)}$ and $\mathbf{S}^{he}$. $\mathbf{S}^{ho(e)} = \{\mathbf{S}^{ho(v)}, \mathbf{S}^{ho(r)}\}$ denotes the affinities within visible and infrared modalities, respectively. 
To enhance the consistency between the homogeneous affinities and the clustering results, we employ the \emph{Jaccard Similarity} \cite{JaccardDist} for clustering to model \emph{\textbf{the homogeneous affinities}}:

\vspace{-2mm}
\begin{equation}\label{intra_sv}
 \mathbf{S}^{ho(v)}_{ij} = \frac{|\mathcal{R}(\mathbf{f}^v_i, \kappa) \cap \mathcal{R}(\mathbf{f}^v_j, \kappa)|}{|\mathcal{R}(\mathbf{f}^v_i, \kappa) \cup \mathcal{R}(\mathbf{f}^v_j, \kappa)|}, \mathbf{S}^{ho(v)} \in \mathbb{R}^{N^v \times N^v},
\end{equation}

\vspace{-2mm}
\begin{equation}\label{intra_sr}
\mathbf{S}^{ho(r)}_{ij} = \frac{|\mathcal{R}(\mathbf{f}^r_i, \kappa) \cap \mathcal{R}(\mathbf{f}^r_j, \kappa)|}{|\mathcal{R}(\mathbf{f}^r_i, \kappa) \cup \mathcal{R}(\mathbf{f}^r_j, \kappa)|}, \mathbf{S}^{ho(r)} \in \mathbb{R}^{N^r \times N^r},
\end{equation}
\vspace{-1mm}


\noindent where $\mathcal{R}(\mathbf{f}_i, \kappa)$ is the $\kappa$-reciprocal nearest neighbors \cite{JaccardDist} of $\mathbf{f}_i$, $i.e.$, $\mathcal{R}(\mathbf{f}_i, \kappa) = \{\mathbf{g_i} | (\mathbf{g_i} \in \mathbb{N}(\mathbf{f}_i, \kappa)) \wedge (\mathbf{f}_i \in \mathbb{N}(\mathbf{g_i}, \kappa)) \}$, and $\mathbb{N}(\mathbf{f}_i, \kappa))$ denotes the $\kappa$-nearest neighbors of the probe $
\mathbf{f}_i$, $|\cdot|$ is the cardinality of a set. 
$\mathbf{S}^{ho}_{ij}$ can be regarded as the affinity between homogeneous features $\mathbf{f}_i$ and $\mathbf{f}_j$.

Due to the substantial modality discrepancy, 
it is inappropriate to directly model the heterogeneous affinity by the \emph{Jaccard Similarities} between cross-modality features as Eq.\ref{intra_sv} and Eq.\ref{intra_sr}.
It should take into account both the cross-modality instance-level relationships and the misalignment between distributions of the two modalities.
Thus affinity modeling can be conceptualized as a transition process between instances from two different distributions, which can be formulated as an optimal transport problem.
Each instance in the dataset should be treated equally, meaning the marginal distributions of instances from both modalities follow uniform distributions.
Thus we model \emph{\textbf{the heterogeneous affinites}} by solving the Optimal Transport (OT) plan according to the transport cost between cross-modality features, which can be formulated as:

\vspace{-2mm}
\begin{equation}\label{OT_problem}
\begin{aligned}
    &\mathop{\min} \limits_{\mathbf{S}^{he}} \langle \mathbf{S}^{he}, \mathbf{C}^{he} \rangle + \frac{1}{\lambda} \langle \mathbf{S}^{he}, -\log(\mathbf{S}^{he}) \rangle. \\
    &s.t. \quad \mathbf{S}^{he} \mathds{1} = \mathds{1} \cdot \frac{1}{N^v}, {\mathbf{S}^{he}}^\top \mathds{1} = \mathds{1} \cdot \frac{1}{N^r},
\end{aligned}
\end{equation}

\noindent 
{where $\langle \cdot \rangle$
denotes the Frobenius dot-product,
and $\mathds{1}$ is an all in 1 vector. } $\mathbf{S}^{he} \in \mathbb{R}^{N^v \times N^r}$ denotes the transport plan. $\mathbf{C}^{he}\in \mathbb{R}^{N^v \times N^r}$ is the cost matrix constructed by computing the Euclidean distance of heterogeneous features, $i.e.$, $\mathbf{C}^{he}_{ij} = \Vert \mathbf{f}^v_i - \mathbf{f}^r_j \Vert^2_2$. 
The first term in the objective function in Eq.\ref{OT_problem} indicates the total transport cost and the second term is the Entropic Regularization \cite{EntropyRegularization}.
The constraints ensure that $N^v$ visible instances are uniformly distributed to $N_r$ infrared instances in the transport plan, and verse vice.
Actually, few instances in one modality inherently exhibit closer proximity to another modality compared to other instances, and such an issue is avoided by the above constraints.
The optimal solution ${\mathbf{S}^{he}}^\ast$ of Eq.\ref{OT_problem} can be obtained by the Sinkhorn-Knopp algorithm \cite{cuturi2013sinkhorn}, and ${\mathbf{S}^{he}_{ij}}^\ast$ can be regarded as the affinity between heterogeneous features $\mathbf{f}^v_i$ and $\mathbf{f}^r_j$. 

We then derive the visible-to-infrared affinity matrix $\mathbf{S}^{he(vr)}={\mathbf{S}^{he}}^\ast\in\mathbb{R}^{N^v \times N^r}$ and the infrared-to-visible affinity matrix $\mathbf{S}^{he(rv)}=
{\mathbf{S}^{he}}^{\ast\top}\in\mathbb{R}^{N^r \times N^v}$
from the heterogeneous affinities ${\mathbf{S}^{he}}^\ast$.
We perform row normalization on the above affinities and obtain the final affinity matrices $\mathbf{S}=\{\mathbf{S}^{ho(v)}, \mathbf{S}^{ho(r)}, \mathbf{S}^{he(vr)}, \mathbf{S}^{he(rv)}\}$, which are subsequently employed to define the structural inconsistency.







\noindent\textbf{Inconsistency Formulation.} 
We leverage the above affinities to define the inconsistency terms for pseudo-labels. A minor degree of structural inconsistency implies that the higher the affinity between two instances, the smaller the distance between their pseudo-labels.
Consequently, we define the product of the Euclidean distance between pseudo-labels of pairwise instances and their affinity as their pairwise structural inconsistency.
From a global perspective, for all instances in both modalities, \emph{\textbf{the homogenous inconsistency}} can be formulated as follows:

\vspace{-2mm}
\begin{equation}\label{Q_intra1}
\mathcal{I}^v_{ho}(\tilde{\mathbf{y}}^{v}) = \sum_{i=1}^{N^v} \sum_{j=1}^{N^v} \mathbf{S}^{ho(v)}_{ij} \Vert {\tilde{\mathbf{y}}}_i^{v} - {\tilde{\mathbf{y}}}_j^{v} \Vert^2_2,
\end{equation}

\vspace{-4mm}
\begin{equation}\label{Q_intra2}
\mathcal{I}^r_{ho}(\hat{\mathbf{y}}^{r}) = \sum_{i=1}^{N^r} \sum_{j=1}^{N^r} \mathbf{S}^{ho(r)}_{ij} \Vert {\hat{\mathbf{y}}}_i^{r} - {\hat{\mathbf{y}}}_j^{r} \Vert^2_2.
\end{equation}
\vspace{-2mm}

\noindent 
The above inconsistency term can be regarded as a weighted sum of pairwise distances between the soft pseudo-labels. The higher the affinities between two instances in feature space, the more their pseudo-label distance contributes to the computation of the inconsistency.
\emph{\textbf{The heterogeneous inconsistency}} can be formulated in a similar manner:


\vspace{-2mm}
\begin{equation}\label{Q_cross1}
\mathcal{I}^v_{he}(\tilde{\mathbf{y}}^v) = \sum_{i=1}^{N^v} \sum_{j=1}^{N^r} \mathbf{S}^{he(vr)}_{ij} \Vert \tilde{\mathbf{y}}_i^v - \hat{\mathbf{y}}_j^r \Vert^2_2,
\end{equation}



Nonetheless, solely minimizing the above two inconsistency terms can lead to a collapse, where all instances end up with nearly identical labels. This is because these two inconsistency statements do not consider the fact that pairwise instances with low affinities should have distinct labels.
Such a reduction in label diversity can negatively impact network training.
To avoid such collapse, a label initialization approach that ensures the diversity of the pseudo-labels is necessary.

Specifically, the intra-modality labels are initialized according to the cluster centroids, and the cross-modality labels are initialized by the widely-used OTLA \cite{OTLA} method.
To fully utilize the relationships between instances and each cluster centroid, we use the soft probability distribution from the corresponding memory bank as initial visible intra-modality labels $\tilde{\mathbf{y}}^v$:
\begin{equation}\label{init_intra}
    \tilde{\mathbf{y}}^v \in \mathbb{R}^{N^v \times K^v},\tilde{\mathbf{y}}^{v}_i(0) = \mathbf{P}(\mathbf{f}_i^v | \tilde{\mathbf{M}}^v, \tau) \in \mathbb{R}^{K^v},
\end{equation}

\noindent where $\mathbf{P}(\mathbf{f} | \mathbf{M}, \tau)$ denotes the probability distribution output from memory bank $\mathbf{M}$ for feature $\mathbf{f}$ with the temperature factor $\tau$, which can be formulated as:

\vspace{-2mm}
\begin{equation}\label{Contrastive_intra}
{{P}_j(\mathbf{f} | \mathbf{M}, \tau)} = \frac{\exp \left(\mathbf{f}^\top \mathbf{c}_{j}/ \tau\right)}{\sum_{k=1}^K \exp \left(\mathbf{f}^\top \mathbf{c}_k / \tau\right)},
\end{equation}

\vspace{-2mm}
\begin{equation}
    \mathbf{P}(\mathbf{f} | \mathbf{M}, \tau) = [P_1(\mathbf{f} | \mathbf{M}, \tau), 
    \cdots, 
    P_{K}(\mathbf{f} | \mathbf{M}, \tau)],
\end{equation}


\noindent {where $K$ denotes the number of prototypes in the memory $\mathbf{M}$,
$\mathbf{c}_j$ and $\mathbf{c}_k$ represents $j$-th and $k$-th cluster prototypes in $\mathbf{M}$.}
Following \cite{OTLA, DOTLA}, we utilize Optimal Transport Label Assignment (OTLA) to initialize infrared cross-modality labels $\hat{\mathbf{y}}^r$:



\vspace{-2mm}
\begin{equation}\label{OTLA}
\begin{aligned}
    &\mathop{\min} \limits_{\mathbf{P}} \langle \mathbf{P}, \mathbf{C} \rangle + \frac{1}{\lambda} \langle \mathbf{P}, -\log(\mathbf{P}) \rangle. \\
    &s.t. \quad \mathbf{P} \mathds{1} = \mathds{1} \cdot \frac{1}{N^r}, \quad \mathbf{P}^\top \mathds{1} = \mathds{1} \cdot \frac{1}{K^v},
\end{aligned}
\end{equation}
\vspace{-2mm}


\noindent 
{where $N^r$ denotes the number of infrared instances, and $K^v$ denotes the number of visible clusters.  $\mathbf{C} \in \mathbb{R}^{N^r \times K^v}$ represents the cost matrix and $\mathbf{P} \in \mathbb{R}^{N^r \times K^v}$ represents the transport plan.}
We utilize the Euclidean distance between infrared instances and visible prototypes from $\tilde{\mathbf{M}}^v$ to calculate the cost, where $\mathbf{C}_{ij} = \Vert \mathbf{f}^r_i - \tilde{\mathbf{m}}^v_j \Vert_2^2$. 
$\mathbf{f}^r_i$ is the $i$-th feature vector in training set and $\tilde{\mathbf{m}}^v_j$ is the j-th prototype in memory $\tilde{\mathbf{M}}^v$.
{The initialized $\hat{\mathbf{y}}^r(0) \in \mathbb{R}^{N^r \times K^v}$ is the one-hot encoding form of the optimal solution $\mathbf{P}^{\ast}$ of Eq.\ref{OTLA}, which is formulated as follows:
}

\vspace{-2mm}
\begin{equation}\label{one_hot_encoding}
{{
\hat{\mathbf{y}}^r_{ik}(0) = 
\left\{ 
\begin{array}{ll}
1, \quad \text{if} \quad k = \arg\max_j \mathbf{P}^{\ast}_{ij} \\
0, \quad \text{otherwise}
\end{array},
\right.
\quad \hat{\mathbf{y}}^r_{i}(0) \in \mathbb{R}^{K^v}.}}
\end{equation}
\vspace{-2mm}

In the optimization process, we aim to ensure that the transferred labels do not deviate too far from the initial labels, thereby guaranteeing diversity in the transferred labels.
Therefore, we propose the \emph{\textbf{self-inconsistency}} to constraint the Euclidean distance between the transferred labels and the initial coarse labels:

\vspace{-2mm}
\begin{equation}\label{Q_self}
\mathcal{I}^v_{self}({\tilde{\mathbf{y}}}^{v}) =\sum_{i=1}^{N^v} \Vert \tilde{\mathbf{y}}_i^{v} - {\tilde{\mathbf{y}}}_{i}^{v}(0) \Vert^2_2.
\end{equation}

\begin{equation}\label{Q_self}
\mathcal{I}^r_{self}({\hat{\mathbf{y}}}^{r}) =\sum_{i=1}^{N^r} \Vert \hat{\mathbf{y}}_i^{r} - {\hat{\mathbf{y}}}_{i}^{r}(0) \Vert^2_2.
\end{equation}

We minimize the above three terms of inconsistency, aiming to derive the pseudo-labels $\tilde{\mathbf{y}}^v$ and $\hat{\mathbf{y}}^r$ with the lowest inconsistency.
This optimization can be formulated as follows:


\vspace{-4mm}
\begin{equation}\label{Q_total_v}
\min \limits_{\tilde{\mathbf{y}}^{v}} \quad \mathcal{I}^e_{ho} (\tilde{\mathbf{y}}^{v}) + \alpha \mathcal{I}^e_{self} (\tilde{\mathbf{y}}^{v}) + (1 - \alpha) \mathcal{I}^e_{he} (\tilde{\mathbf{y}}^{v}), \\ 
\end{equation}

\vspace{-4mm}
\begin{equation}\label{Q_total_r}
\min \limits_{\hat{\mathbf{y}}^{r}} \quad \mathcal{I}^e_{ho} (\hat{\mathbf{y}}^{r}) + \alpha \mathcal{I}^e_{self} (\hat{\mathbf{y}}^{r}) + (1 - \alpha) \mathcal{I}^e_{he} (\hat{\mathbf{y}}^{r}), \\ 
\end{equation}

\noindent where $\alpha$ is a trade-off parameter.

\noindent \textbf{Label Transfer.} The above optimization problem can be solved by the LGC algorithm \cite{LGC}. 
Specifically, the visible and infrared modality-unified labels are updated alternately as Eq.\ref{v_update} and Eq.\ref{r_update} until convergence:

\vspace{-4mm}
\begin{equation}\label{v_update}
\begin{array}{lc}
\vspace{1ex}
\tilde{\mathbf{y}}^{v}(t+1)^{\ast} = (1 - \alpha) \mathbf{S}^{he(vr)} \cdot \hat{\mathbf{y}}^{r}(t) + \alpha \tilde{\mathbf{y}}^{v}(0),
\\
\tilde{\mathbf{y}}^{v}(t+1) = \frac{1}{2} \mathbf{S}^{ho(v)} \cdot \tilde{\mathbf{y}}^{v}(t+1)^{\ast}+ \frac{1}{2} \tilde{\mathbf{y}}^{v}(t+1)^{\ast},
\end{array}
\end{equation}

\vspace{-4mm}
\begin{equation}\label{r_update}
\begin{array}{lc}
\vspace{1ex}
\hat{\mathbf{y}}^{r}(t+1)^{\ast} = (1 - \alpha) \mathbf{S}^{he(rv)} \cdot \tilde{\mathbf{y}}^{v}(t) + \alpha \hat{\mathbf{y}}^{r}(0),
\\
\vspace{1ex}
\hat{\mathbf{y}}^{r}(t+1) = \frac{1}{2} \mathbf{S}^{ho(r)} \cdot \hat{\mathbf{y}}^{r}(t+1)^{\ast}+ \frac{1}{2} \hat{\mathbf{y}}^{r}(t+1)^{\ast},
\end{array}
\vspace{-2mm}
\end{equation}

\noindent 
where $\mathbf{y}^e(t) = \{\tilde{\mathbf{y}}^v(t), \hat{\mathbf{y}}^r(t)\}$ denotes the labels in $t-th$ iteration during the optimization process. 
$\mathbf{y}^e(t+1)^{\ast}$ denotes the intermediate variables during the update process in one iteration.
The upper terms in Eq.\ref{v_update} and Eq.\ref{r_update} can be regarded as a cross-modality transfer process, while the lower terms stand for a manner to further compensate the transferred pseudo-labels utilizing intra-modality structural information.

{In detail, take the pseudo-labels $\tilde{\mathbf{y}}^v$ in visible modality, the inconsistency formulation Eq.\ref{Q_total_v} can be expressed in the form of vectors and matrices:}

\begin{equation}
{
\begin{aligned}
\mathcal{I} (\tilde{\mathbf{y}}^v)
&=\alpha \mathbf{tr} ((\tilde{\mathbf{y}}^v - \tilde{\mathbf{y}}^v(0))^\top
(\tilde{\mathbf{y}}^v - \tilde{\mathbf{y}}^v(0))) \\
&+ (1 - \alpha) \mathbf{tr}
((\tilde{\mathbf{y}}^v - \mathbf{\mathbf{S}}^{he(vr)} \cdot \hat{\mathbf{y}}^r)^\top
(\tilde{\mathbf{y}}^v - \mathbf{\mathbf{S}}^{he(vr)} \cdot \hat{\mathbf{y}}^r)) \\
&+ \mathbf{tr}(\tilde{\mathbf{y}}^{v\top} (\mathbf{I}_{N^v} - \mathbf{S}^{ho(v)}) {\tilde{\mathbf{y}}^v}),
\end{aligned}
}
\vspace{0mm}
\end{equation}


\noindent {
where $\mathbf{I}_{N^v}$ denotes the $N^v \times N^v$ identity matrix and $\mathbf{tr} (\cdot)$ denotes the trace of matrix.
To minimize the inconsistency $\mathcal{I} (\tilde{\mathbf{y}}^v)$ for visible pseudo-labels, we compute its partial derivative with respect to $\tilde{\mathbf{y}}^v$ and equate this partial derivative to zero, where $\frac{\partial{\mathcal{I} (\tilde{\mathbf{y}}^v)}}{\partial{\tilde{\mathbf{y}}^v}} = 0$
Then $\tilde{\mathbf{y}}^v$ can be represented as:
}

\begin{equation}\label{ditui_supp}
{
\tilde{\mathbf{y}}^v = \frac{1}{2} \mathbf{S}^{ho(v)} \cdot \tilde{\mathbf{y}}^v + 
\frac{1}{2} \left[
(1 - \alpha)
\mathbf{S}^{he(vr)} \cdot \hat{\mathbf{y}}^r +
\alpha \tilde{\mathbf{y}}^v(0) \right].
}
\vspace{0mm}
\end{equation}

To enhance its convergence stability, we decompose Eq.\ref{ditui_supp} into two steps, $i.e.$, cross-modality transfer, and intra-modality transfer, which is formulated as follows:

\begin{equation}
\begin{aligned}
&\tilde{\mathbf{y}}^v
\xleftarrow{\text{cross-modality transfer}} (1 - \alpha) \mathbf{S}^{he(vr)} \cdot \hat{\mathbf{y}}^r + \alpha \tilde{\mathbf{y}}^v(0), \\
&\tilde{\mathbf{y}}^v \xleftarrow{\text{intra-modality transfer}} \frac{1}{2} \mathbf{S}^{ho(v)} \cdot \tilde{\mathbf{y}}^v + \frac{1}{2} \tilde{\mathbf{y}}^v.
\end{aligned}
\vspace{0mm}
\end{equation}

\noindent{During each iterative optimization process, the pseudo-labels $\tilde{\mathbf{y}}^v$ are updated by alternating iterations of fixed points, which is formulated as Eq.\ref{v_update}.}

\begin{algorithm}[t] 
  \caption{V2R MULT}  
  \label{alg::MULT}  
    \KwIn{
      $\mathbf{f}^v$: Visible features;
      $\mathbf{f}^r$: Infrared features;}
    \KwOut{\\Visible intra-modality labels $\tilde{\mathbf{y}}^v$;\\
    {Infrared cross-modality labels $\hat{\mathbf{y}}^r$};}
    \textbf{Initialization:\\} Two intra-modality memory banks $\tilde{\mathbf{M}}^v$ and $\tilde{\mathbf{M}}^r$\;
    Intra-modality visible labels $\tilde{\mathbf{y}}^v(0)$ (Eq.\ref{init_intra})\;
    Cross-modality infrared labels  $\hat{\mathbf{y}}^r(0)$ (Eq.\ref{OTLA})\;
    \textbf{Affinity Modeling:\\} 
    {Intra-modality affinity matrices $\mathbf{S}^{v}$ and $\mathbf{S}^r$\\
    (Eq.\ref{intra_sv} \& Eq.\ref{intra_sr})}\;
    Cross-modality affinity matrices $\mathbf{S}^{vr}$ and $\mathbf{S}_{rv}$ (Eq.\ref{OT_problem})\;
    \textbf{Do Label Transfer (V2R):\\}
    $t = 0$; $\epsilon_0 = 1e-2$; $\epsilon=1e6$\;
    \While{$\epsilon > \epsilon_0$}
    {Calculate $\tilde{\mathbf{y}}^v(t+1)$ according to Eq.\ref{v_update}\;
    Calculate $\hat{\mathbf{y}}^r(t+1)$ according to Eq.\ref{r_update}\;
    $\epsilon \leftarrow$ \\
    $\max (\Vert \tilde{\mathbf{y}}^v(t+1) - \tilde{\mathbf{y}}^v(t) \Vert_1, \Vert \hat{\mathbf{y}}^r(t+1) - \hat{\mathbf{y}}^r(t) \Vert_1)$\;
    $t \leftarrow t + 1$\;
    }
    Update labels $\tilde{\mathbf{y}}^v \leftarrow \tilde{\mathbf{y}}^v(t)$; $\hat{\mathbf{y}}^r \leftarrow \hat{\mathbf{y}}^r(t)$\;
    \textbf{Return:} Soft labels $\tilde{\mathbf{y}}^v$ and $\hat{\mathbf{y}}^r$ (Eq.\ref{soft}).
\end{algorithm}

Nevertheless, the transferred pseudo-labels are excessively smoothed, which is detrimental to the network's ability to learn predictions with low entropy. Therefore, we derive the soft modality-unified labels by aggregating both hard labels and soft forms of the pseudo-labels, which can be formulated as:

\vspace{-2mm}
\begin{equation}\label{soft}
    \mathbf{y}^e = \beta \cdot \mathbf{y}_{hard}^e + (1 - \beta) \cdot \mathbf{y}_{soft}^e,
\end{equation}
\vspace{-2mm}

\noindent where $\beta$ is a parameter to control the smoothness of the pseudo-label. $\mathbf{y}^e_{hard}$ and $\mathbf{y}^e_{soft}$ denotes the hard one-hot encoding pseudo-labels and the soft pseudo-labels obtained from the transferred labels $\mathbf{y}^e = \{ \tilde{\mathbf{y}}^v, \hat{\mathbf{y}}^r\}$, respectively.
{The one-hot form $\mathbf{y}^e_{hard}$ is directly derived from the soft label $\mathbf{y}^e_{soft}$.}

It is noteworthy that, the labels $\tilde{\mathbf{y}}_i^v$ and $\hat{\mathbf{y}}_i^r$ have a same dimension of $K^v$. 
The detailed algorithm flow of our V2R MULT is in Alg.\ref{alg::MULT}. The R2V MULT, which provides $\tilde{\mathbf{y}}^r$ and $\hat{\mathbf{y}}^v$ in infrared label space, holds a symmetrical form with V2R MULT.

\subsection{Alternative Modality-Invariant Representation Learning}\label{MIRL}

Guided by the modality-unified soft labels from the MULT module, we lay emphasis on alleviating the cross-modality discrepancy while keeping intra-modality consistency. 
We concurrently execute intra-modality and cross-modality contrastive learning. 
Additionally, we design an alternative scheme that alternates between V-based and R-based training modes.

\begin{figure}
\centering
\includegraphics[width=0.48\textwidth]{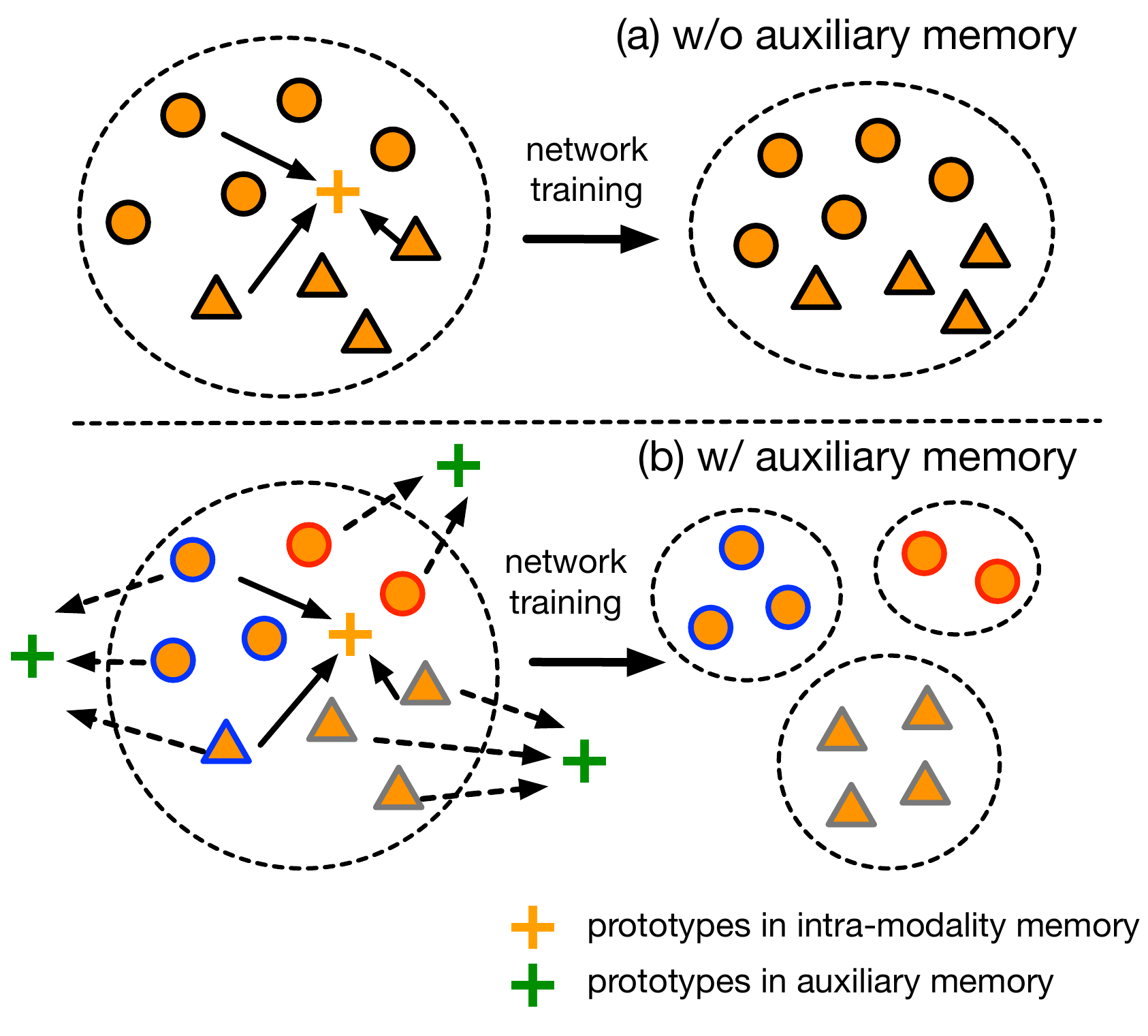}
\caption{
Illustration of the role of the auxiliary memory. 
Different shapes denote different identities and
different edge colors denote different cross-modality labels.
}\label{aux_motivation}
\vspace{-3mm}
\end{figure}

To be specific, for V-based training mode, we construct a cross-modality memory bank $\mathbf{M}^c \in \mathbb{R}^{K^v \times d}$ and an auxiliary memory bank $\hat{\mathbf{M}}^r \in \mathbb{R}^{K^v \times d}$, which are initialized in the same manner as the visible intra-modality memory bank $\tilde{\mathbf{M}}^v$. 
$\mathbf{M}^c$ is updated by instances from both modalities and $\hat{\mathbf{M}}^r$ is updated by instances only from the infrared modality.
The auxiliary memory bank $\hat{\mathbf{M}}^r$ aims to learn the infrared intra-modality structure depicted by its labels $\hat{\mathbf{y}}^r$ in visible label space, thereby serving as a compensatory mechanism for $\tilde{\mathbf{M}}^r$.
In fact, instances from different identities are sometimes gathered in the same cluster with their intra-modality labels ($\tilde{\mathbf{y}}^r$), while they are distributed correctly into different clusters with their cross-modality labels ($\hat{\mathbf{y}}^r$).
Thus the auxiliary memory, which reflects the structure of $\hat{\mathbf{y}}^r$, can assist the intra-modality learning.
As shown in Flg.\ref{aux_motivation}, compared to utilizing prototypes solely from the intra-modality memory, leveraging prototypes from the auxiliary memory of another modality is advantageous for uncovering fine-grained structural details within clusters in both two modalities.  

Following the widely-used memory-based methods \cite{SPCL, ClusterContrast, ADCA, memory-based-VI-ReID}, we carry out contrastive learning during forward-propagation (FP) and update the memory banks during backward-propagation (BP). The visible intra-modality contrastive loss can be formulated as: 

\vspace{-2mm}
\begin{equation}\label{L_IM_v}
    \mathcal{L}^v_{IM} = \frac{1}{B} \sum^B_{i=1} \mathcal{L}_{ce}(\mathbf{P}(\mathbf{f}_i^v | \tilde{\mathbf{M}}^v, \tau),\tilde{\mathbf{y}}^v_i),
\end{equation}


\noindent {where $\mathcal{L}_{ce}$ denotes the standard cross-entropy loss and $B$ denotes the input batch size.} For infrared modality in V-based training, both $\tilde{\mathbf{y}}^r$ and $\hat{\mathbf{y}}^r$ are utilized jointly during contrastive learning with their corresponding memory $\tilde{\mathbf{M}}^r$ and $\hat{\mathbf{M}}^r$:

\vspace{-2mm}
\begin{equation}\label{L_IM_r}
\begin{aligned}
    \mathcal{L}^r_{IM} & = \frac{1}{B} \sum^B_{i=1} \mathcal{L}_{ce}(\mathbf{P}(\mathbf{f}_i^r | \tilde{\mathbf{M}}^r, \tau), \tilde{\mathbf{y}}^r_i) \\
    & + \frac{1}{B} \sum^B_{i=1}
    \mathcal{L}_{ce}(\mathbf{P}(\mathbf{f}_i^r | \hat{\mathbf{M}}^r, \tau), \hat{\mathbf{y}}^r_i).
\end{aligned}
\end{equation}
\vspace{-2mm}


\noindent 
These two types of labels in two different label spaces play a mutual corrective role and facilitate intra-modality training. The total intra-modality contrastive loss can be formulated as follows:

\vspace{-4mm}
\begin{equation}
    \mathcal{L}_{IM} = \mathcal{L}^v_{IM} + \mathcal{L}^r_{IM}.
\end{equation}
\vspace{-4mm}

Besides, we perform cross-modality contrastive learning to constrain the relationships between instances in different modalities and their corresponding modality-agnostic cluster prototypes in $\mathbf{M}^c$ in the feature space:

\vspace{-4mm}
\begin{equation}\label{L_CM}
\begin{aligned}
    \mathcal{L}_{CM}
    & = \frac{1}{B} \sum^B_{i=1}
    \mathcal{L}_{ce}(\mathbf{P}(\mathbf{f}_i^v | \mathbf{M}^c, \tau), \tilde{\mathbf{y}}^v_i) \\
    & + \frac{1}{B} \sum^B_{i=1}
    \mathcal{L}_{ce}(P(\mathbf{f}_i^r | \mathbf{M}^c, \tau), \hat{\mathbf{y}}^r_i).
\end{aligned}
\end{equation}

During the BP stage, the memory banks are updated with the input features in a momentum strategy \cite{moco, ClusterContrast}:


\vspace{-4mm}
\begin{equation}
\begin{aligned}
& \tilde{\mathbf{m}}^v(\tilde{y}_i^v) \leftarrow \mu \tilde{\mathbf{m}}^v(\tilde{y}_i^v) + (1 - \mu) \mathbf{f}(\tilde{y}_i^v), \mathbf{f} \in \{\mathbf{f}_1^v, \cdots, \mathbf{f}_B^v\} \\
& \tilde{\mathbf{m}}^r(\tilde{y}_i^r) \leftarrow \mu \tilde{\mathbf{m}}^v(\tilde{y}_i^r) + (1 - \mu) \mathbf{f}(\tilde{y}_i^r), \mathbf{f} \in \{\mathbf{f}_1^r, \cdots, \mathbf{f}_B^r\} \\
& \mathbf{m}^c(\tilde{y}_i^v) \leftarrow \mu \mathbf{m}^c(\tilde{y}_i^v) + (1 - \mu) \mathbf{f}(\tilde{y}_i^v), \mathbf{f} \in \{\mathbf{f}_1^v, \cdots, \mathbf{f}_B^v\} \\
& \mathbf{m}^c(\hat{y}_i^r) \leftarrow \mu \mathbf{m}^c(\hat{y}_i^r) + (1 - \mu) \mathbf{f}(\hat{y}_i^r), \mathbf{f} \in \{\mathbf{f}_1^r, \cdots, \mathbf{f}_B^r\} \\
& \hat{\mathbf{m}}^r(\hat{y}_i^r) \leftarrow \mu \hat{\mathbf{m}}^r(\hat{y}_i^r) + (1 - \mu) \mathbf{f}(\hat{y}_i^r), \mathbf{f} \in \{\mathbf{f}_1^r, \cdots, \mathbf{f}_B^r\}, \\
\end{aligned}
\end{equation}

\noindent where $\mu$ is the momentum updating factor. $\mathbf{m}(y)$ is the $y-th$ prototype in the memory bank $\mathbf{M}$, and $\mathbf{f}(y)$ is the input feature with label $y$ in current mini-batch. 
Note that $y = \{\tilde{y}_i^v, \tilde{y}_i^r, \hat{y}_i^r\}$ for memory update is the hard label form of the soft labels $\mathbf{y} = \{\tilde{\mathbf{y}}_i^v, \tilde{\mathbf{y}}_i^r, \hat{\mathbf{y}}_i^r\}$.

{
For R-based training, a cross-modality memory bank $\mathbf{M}^c \in \mathbb{R}^{K^r \times d}$ is constructed according to labels $\tilde{\mathbf{y}}^r$ and $\hat{\mathbf{y}}^v$ from R2V MULT. Meanwhile an auxiliary memory bank $\hat{\mathbf{M^v}} \in \mathbb{R}^{K^r \times d}$ is initialized in the same manner as $\tilde{\mathbf{M}}^r$.
It is updated by visible features based on $\hat{\mathbf{y}}^v$ in the infrared label space. The loss functions for R-based training can be written in a symmetric form in comparison to the V-based training. }

{The entire training process is described in Alg.\ref{alg::training}. We conduct an alternate scheme for the V-based training and the R-based training as the epoch goes on: when epoch-id $E\%2==0$, the V-based training is conducted; otherwise, the R-based training is conducted.}
In one epoch, the model is supervised by labels from MULT of a specific direction.
In adjacent epochs, the model is supervised by labels from MULT of different directions.
Such an alternative scheme mitigates the inconsistency between pseudo-labels from MULT of different directions (V2R / R2V).

\begin{algorithm}[t] 
  \caption{Training process in one epoch.}  
  \label{alg::training}  

    {\textbf{Input:} epoch $E$; network $f_{\theta}$; iteration number from $T_0$ to $T_t$ in one epoch; batch size $B$;\\}
    1: Extract visible features $\left\{\mathbf{f}^v_i | i=1, 2, \cdots, N^v\right\}$ and infrared features $\left\{\mathbf{f}^r_i | i=1, 2, \cdots, N^r\right\}$;\\
    2: Clustering features and initialize two intra-modality memory banks $\tilde{\mathbf{M}}^v$ and $\tilde{\mathbf{M}}^r$ based on the cluster centroids; \\
    3: Do V2R MULT / R2V MULT according to Alg.\ref{alg::MULT}; \\
    \uIf{$E \% 2 == 0:$}{
    1: Initialize cross-modality memory $\mathbf{M}^c$ and auxiliary memory $\hat{\mathbf{M}}^r$ according to the results from V2R MULT; \\
    2: \For{$t = T_0$ to $T_t$}
    {
    Do V-based training;
    }}
    \ElseIf{$E \% 2 == 1$}{
    1: Initialize cross-modality memory $\mathbf{M}^c$ and auxiliary memory $\hat{\mathbf{M}}^v$ according to the results from R2V MULT; \\
    2: \For{$t = T_0$ to $T_t$}
    {
    Do R-based training;
    }}
    4: \textbf{Output:} updated network $f_{\theta}.$  
\end{algorithm}

\begin{table*}[!htbp]
    \centering	
    \caption{Comparison with the state-of-the-art methods on SYSU-MM01 and RegDB. ``GUR$^\ast$" denotes GUR without camera labels. 
    Since our method does not require any camera label information, for fair comparison we do not report the results of GUR with camera labels.}
    \begin{adjustbox}{max width=\textwidth}
    \footnotesize
    \begin{tabular}{c|c|ccc|ccc|ccc|ccc}
    \hline
        \multicolumn{1}{c|}{
        \multirow{3}{*}{Method}} &  \multicolumn{1}{c|}{
        \multirow{3}{*}{Venue}} &
    \multicolumn{6}{c|}{SYSU-MM01 (Single-shot)} & 
        \multicolumn{6}{c}{RegDB}\\
        \cline{3-14}
        & & \multicolumn{3}{c|}{All-search} & \multicolumn{3}{c|}{Indoor-search} &
        \multicolumn{3}{c|}{Visible-to-Infrared} & \multicolumn{3}{c}{Visible-to-Infrared}\\ 
        \cline{3-14}
    & & R1 & mAP & mINP & R1 & mAP & mINP & R1 & mAP & mINP &R1 & mAP & mINP\\
        \hline
        \multicolumn{12}{l}{\textit{Supervised VI-ReID methods}} \\ \hline
        Zero-Pad\cite{SYSU-MM01} & ICCV-17 & 14.80 & 15.95 & - & 20.58 & 26.92 & - & 17.75 & 18.90 & - & 16.63 & 17.82 & -\\
        AlignGAN\cite{AlignGAN} & ICCV-19 & 42.4 & 40.7 & - & 45.9 & 54.3 & - & 57.9 & 53.6 & - & 56.3 & 53.4 & - \\
        cm-SSFT \cite{cm-SSFT} & CVPR-20 & 47.7 & 54.1 & - & - & - & - & 72.3 & 72.9 & - & 71.0 & 71.7 & - \\
        DDAG \cite{DDAG} & ECCV-21 & 54.75 & 53.02 & 39.62 & 61.02 & 67.98 & 62.61 & 69.34 & 63.46 & 49.24 & 68.06 & 61.80 & 48.62 \\
        AGW \cite{agw} & TPAMI-21 & 47.50 & 47.65 & 35.30 & 54.17 & 62.97 & 59.23 & 70.05 & 66.37 & 50.19 & 70.49 & 65.90 & 51.24\\
        VCD+VML \cite{VCD+VML} & CVPR-21 & 60.02 & 58.80 & - & 66.05 & 72.98 & - & 73.2 & 71.6 & - & 71.8 & 70.1 & - \\
       CA \cite{CA} & ICCV-21 & 69.88 & 66.89 & 53.61 & 76.26 & 80.37 & 76.79 & 85.03 & 79.14 & 65.33 & 84.75 & 77.82 & 61.56 \\
        MPANet \cite{MPANet} & CVPR-21 & 70.58 & 68.24 & - & 76.74 & 80.95 & - & 82.8 & 80.7 & - & 83.7 & 80.9 & -\\
        MAUM \cite{MAUM} & CVPR-22 & 71.68 & 68.79 & - & 76.97 & 81.94 & - & 87.87 & 85.09 & - & 86.95 & 84.34 & -\\
        CIFT \cite{CIFT} & ECCV-22 & 74.08 & \underline{74.79} & - & \underline{81.82} & \underline{85.61} & - & \underline{91.96} & \underline{92.00} & - & 90.30 & \underline{90.78} & - \\
        DEEN\cite{DEEN} & CVPR-23 & 74.7 & 71.8 & - & 80.3 & 83.3 & - & 91.1 & 85.1 & - & 89.5 & 83.4 & - \\
        SEFEL \cite{SEFEL} & CVPR-23 & 77.12 & 72.33 & - & 82.07 & 82.95 & - & 91.07 & 85.23 & - & \underline{92.18} & 86.59 & - \\
        PartMix \cite{PartMix} & CVPR-23 & \underline{77.78} & 74.62 & - & 81.52 & 84.38 & - & 84.93 & 82.52 & - & 85.66 & 82.27 & - \\
        \hline
        \multicolumn{12}{l}{\textit{Unsupervised single-modality ReID methods}} \\ \hline
        SPCL \cite{SPCL} & NIPS-20 & 18.37 & 19.39 & 10.99 & 26.83 & 36.42 & 33.05 & 13.59 & 14.86 & 10.36 & 11.70 & 13.56 & 10.09 \\
        MMT \cite{MMT} & ICLR-20 & \underline{21.47} & 21.53 & 11.50 & 22.79 & 31.50 & 27.66 & \underline{25.68} & \underline{26.51} & \underline{19.56} & \underline{24.42} & \underline{25.59} & \underline{18.66} \\
        Cluster-Contrast \cite{ClusterContrast} & arXiv-21 & 20.16 & \underline{22.00} & \underline{12.97} & 23.33 & 34.01 & 30.88 & 11.76 & 13.88 & 9.94 & 11.14 & 12.99 & 8.99 \\
        ICE \cite{ICE} & ICCV-21 & 20.54 & 20.39 & 10.24 & \underline{29.81} & \underline{38.35} & \underline{34.32} & 12.98 & 15.64 & 11.91 & 12.18 & 14.82 & 10.6 \\
        PPLR \cite{PPLR} & CVPR-22 & 11.98 & 12.25 & 4.97 & 12.71 & 20.81 & 17.61 & 10.30 & 11.94 & 8.10 & 10.39 & 11.23 & 7.04 \\
        ISE \cite{ISE} & CVPR-22 & 20.21 & 18.93 & 8.54 & 14.22 & 24.62 & 21.74 & 16.12 & 16.99 & 13.24 & 10.83 & 13.66 & 10.71 \\
        \hline
        \multicolumn{12}{l}
        {\textit{Unsupervised VI-ReID methods}} \\ \hline
        H2H \cite{H2H} & TIP-21 & 25.49 & 25.16 & - & - & - & - & 13.91 & 12.72 & - & 14.11 & 12.29 & - \\
        H2H(AGW) \cite{H2H} & TIP-21 & 30.15 & 29.40 & - & - & - & - & 23.81 & 18.87 & - & - & - & - \\
        H2H(AGW) w/ CMRR \cite{H2H} & TIP-21 & 45.47 & 47.99 & - & - & - & - & 35.18 & 36.46 & - & - & - & - \\
        OTLA \cite{OTLA} & ECCV-22 & 29.9 & 27.1 & - & 29.8 & 38.8 & - & 32.9 & 29.7 & - & 32.1 & 28.6 & - \\
        ADCA \cite{ADCA} & MM-22 & 45.51 & 42.73 & 28.29 & 50.60 & 59.11 & 55.17 & 67.20 & 64.05 & 52.67 & 68.48 & 63.81 & 49.62 \\
        ADCA(AGW) \cite{ADCA} & MM-22 & 50.90 &  45.70 & 29.12 & 51.39 & 59.82 & 56.08 & 66.62 & 63.47 & - & 67.29 & 62.98 & -\\
        CHCR (AGW) \cite{CHCR} & T-CSVT-23 & 47.72 & 45.34 & - & - & - & - & 68.18 & 63.75 & - & 69.96 & 65.87 & - \\ 
        DOTLA (AGW) \cite{DOTLA} & MM-23 & 50.36 & 47.36 & 32.40 & 53.47 & 61.73 & 57.35 & \underline{85.63} & 76.71 & 61.58 & \underline{82.91} & 74.97 & 58.60 \\
        MBCCM \cite{MBCCM} & MM-23 & 53.14 & 48.16 & 32.41 & 55.21 & 61.98 & 57.23 & 83.79 & \underline{77.87} & \underline{65.04} & 82.82 & \underline{76.74} & \underline{61.73} \\
        MIMR \cite{MIMR} & KBS-24 & 46.56 & 45.88 & - & 52.26 & 60.93 & - & 68.76 & 64.33 & - & 68.76 & 63.83 & -\\
        CCLNet (CLIP) \cite{CCLNet} & MM-23 & 
        54.03 & 50.19 & - & 56.68 & 65.12 & - &69.94 & 65.53 & - & 70.17 & 66.66 & -\\
        PGM(AGW) \cite{PGM} & CVPR-23 & 57.27 & 51.78 & 34.96 & 56.23 & 62.74 & 58.13 & 69.48 & 65.41 & - & 69.85 & 65.17 & - \\
        GUR$^\ast$ \cite{GRU} & ICCV-23 &
        \underline{60.95} & \underline{56.99} & \underline{41.85} & \underline{64.22} & \underline{69.49} & \underline{64.81} & 73.91 & 70.23 & 58.88 & 75.00 & 69.94 & 56.21 \\
        \hline
    Ours & - & \textbf{64.77} & \textbf{59.23} & \textbf{43.46} & \textbf{65.34} & \textbf{71.46} & \textbf{67.83} & \textbf{89.95} & \textbf{82.09} & \textbf{67.29} & \textbf{90.78} & \textbf{82.25} & \textbf{65.38} \\
    Ours w/ CMRR \cite{H2H} & Ours + TIP-21 & \textbf{71.23} & \textbf{67.54} & \textbf{54.48} & \textbf{69.06} & \textbf{75.06} & \textbf{71.36} & \textbf{91.50} & \textbf{92.51} & \textbf{92.03} & \textbf{90.53} & \textbf{91.49} & \textbf{91.27} \\
    \hline	
    \end{tabular}
    \end{adjustbox}
    \vspace{-3mm}
\label{tab:sota_comparison}
\end{table*}

\subsection{Online Cross-memory Label Refinement}\label{OCLR}

While the MULT module enables training under modality-unified supervision, it is essential to consider that due to the inherent noise in clustering and MULT, prolonged training with fixed supervised signals may cause the network to overfit incorrect labels.


To overcome this issue, we utilize the predictions from the intra-modality memory banks to refine the predictions from the cross-modality memory bank. 
Several reasons support the adoption of such refinement strategies:
(1) The intra-modality memory banks are updated by features within their respective modalities, this update occurs at a slower rate compared to the cross-modality memory bank, resulting in more stable predictions.
(2) The intra-modality and cross-modality memory banks are expected to output consistent predictions.
{In the visible modality, we adopt mutual refinement between the extracted features $\mathbf{f}_i^v$ from visible images and $\mathbf{f}_i^a$ from images with CA augmentation \cite{CA} to enhance constraints on their self-consistency.
CA is a widely-used data augmentation strategy in VI-ReID \cite{ADCA, MBCCM, PGM}, which randomly selects one channel in a visible image and expands it into a three-channel gray image. 
The augmented images build an intermediate modality between visible and infrared to enhance training.
}
The OCLR loss for visible modality in V-based training mode can be formulated as follows:

\vspace{-4mm}
\begin{equation}\label{L_OCLR}
\vspace{0mm}
\begin{aligned}
    \mathcal{L}^v_{OCLR}
    & = \frac{1}{B} \sum^B_{i=1}    \mathcal{L}_{ce}(\mathbf{P}(\mathbf{f}_i^v | \mathbf{M}^c, \tau), \mathbf{P}(\mathbf{f}_i^a | \tilde{\mathbf{M}}^v, \frac{1}{5} \tau)) \\
    & + \frac{1}{B} \sum^B_{i=1}
    \mathcal{L}_{ce}(\mathbf{P}(\mathbf{f}_i^a | \mathbf{M}^c, \tau), \mathbf{P}(\mathbf{f}_i^v | \hat{\mathbf{M}}^r, \frac{1}{5} \tau)).
\end{aligned}
\end{equation}
\vspace{-2mm}

\noindent 
As in \cite{DINO, protocon}, we sharpen the target prediction more than the source's to encourage entropy minimization. The setting of the sharpen ratio follows ProtoCon \cite{protocon}. The OCLR loss for infrared modality $\mathcal{L}^r_{OCLR}$ in V-based mode can be formulated similarly:

\vspace{-4mm}
\begin{equation}\label{L_OCLR}
\vspace{0mm}
\begin{aligned}
    \mathcal{L}^r_{OCLR}
    & = \frac{1}{B} \sum^B_{i=1}
    \mathcal{L}_{ce}(\mathbf{P}(\mathbf{f}_i^r | \mathbf{M}^c, \tau), \mathbf{P}(\mathbf{f}_i^r | \tilde{\mathbf{M}}^v, \frac{1}{5} \tau)) \\
    & + \frac{1}{B} \sum^B_{i=1}
    \mathcal{L}_{ce}(\mathbf{P}(\mathbf{f}_i^r | \mathbf{M}^c, \tau), \mathbf{P}(\mathbf{f}_i^r | \hat{\mathbf{M}}^r, \frac{1}{5} \tau)).
\end{aligned}
\end{equation}
\vspace{-6mm}

\subsection{The Overall Objective Function}

The overall training loss can be formulated as follows:

\vspace{-4mm}
\begin{equation}\label{L_total}
    \mathcal{L} = \mathcal{L}_{IM} + \mathcal{L}_{CM} + \mathcal{L}^v_{OCLR} + \mathcal{L}^r_{OCLR},
    \vspace{-1mm}
\end{equation}
\vspace{-4mm}


\noindent where $\mathcal{L}_{IM}$, $\mathcal{L}_{CM}$, $\mathcal{L}^v_{OCLR}$ and $\mathcal{L}^r_{OCLR}$ are described in detail in Sec.\ref{MIRL} and Sec.\ref{OCLR}. 

\section{Experiments}

\subsection{Dataset and Evaluation Protocol}

\textbf{Datasets.} Our proposed method is evaluated on two public visible-infrared datasets, namely SYSU-MM01 \cite{SYSU-MM01} and RegDB \cite{RegDB}. 
SYSU-MM01 comprises 395 identities, with 22258 visible and 11909 infrared images captured by indoor and outdoor cameras.
The dataset is divided into the training set (296 IDs), the validation set (99 IDs), and the test set (96 IDs). 
RegDB is a smaller dataset obtained from a pair of aligned visible and infrared cameras \cite{RegDB}. It contains 412 identities, where each identity has 10 visible images and 10 infrared images.   

\textbf{Evaluation Metrics.} All experiments follow the common evaluation protocols used for VI-ReID \cite{agw, DDAG, CA}. 
The evaluation metrics include Cumulative Matching Characteristic (CMC), Mean Average Precision (mAP), and Mean Inverse Negative Penalty (mINP \cite{agw}).
For the SYSU-MM01 dataset, 
following \cite{MPANet, cm-SSFT, AlignGAN}, we evaluate the proposed method under two search modes: the All-search mode and the Indoor-search mode.
In All-search mode, the gallery set contains all visible images from both indoor and outdoor cameras, while in
indoor-search mode, the gallery set only contains images from indoor cameras.
Additionally,
we conduct experiments under both single-shot and multi-shot settings.
We repeat the evaluation 10 times with the random split of the query set and the gallery set and report the average performance.
As for the RegDB dataset, we evaluate our method on two testing modes: Visible-to-Infrared and Infrared-to-Visible. 
The dataset is randomly split into training (206 identities) and testing (206 identities) sets.
We randomly selected 206 identities for training and the remaining 206 for testing. 

\textbf{Implementation Details.} 
Our proposed method is implemented using PyTorch.
we adopt two-stream ResNet50 \cite{resnet} pretrained on ImageNet \cite{ImageNet} as our backbone. 
In a mini-batch, the number of classes $P$ and instances for each class $K$ are both 12. All the input images are resized to $288 \times 144$.
The augmentations for images are following \cite{ADCA}. Besides, Linear Transform Generator (LTG \cite{LTG}) and ColorJitter augmentations are adopted in the DCL training stage to promote the encoder to extract color-independent features. 
We train for a total of 90 epochs.
The DCL framework is trained in the first 40 epochs, while our proposed framework is trained in the other 50 epochs.
We use the Adam optimizer for training the model with weight decay 5e-4. The initial learning rate is set to 3.5e-4 and decays 10 times at the 20th, 60th, and 80th epochs.
Following \cite{MBCCM, ADCA, PGM}, the momentum factor $\mu$ is 0.1, and the temperature $\tau$ is 0.05, while the maximum distance for DBSCAN is set to 0.6 on SYSU-MM01 and 0.3 on RegDB. 
The parameter $\kappa$ for $\kappa$-reciprocal nearest neighbors in Eq.\ref{intra_sv} and Eq.\ref{intra_sr} is set to 30 following \cite{JaccardDist}.
The hyper-parameter $\lambda$ in Eq.\ref{OTLA} for OT is set to 25. 
The trade-off parameter $\alpha$ in Eq.\ref{Q_total_v} and Eq.\ref{Q_total_r} is set to 0.2 and $\beta$ in Eq.\ref{soft} is set to 0.7.

\begin{table}[!htbp]
    \centering	
    \caption{Comparison with the state-of-the-art methods under the Multi-shot setting on SYSU-MM01.}
    \begin{adjustbox}{max width=0.48\textwidth}
    \footnotesize
    \begin{tabular}{c|cc|cc}
    \hline
        \multicolumn{1}{c|}{
        \multirow{3}{*}{Method}} &
    \multicolumn{4}{c}{SYSU-MM01 (Multi-shot)} \\
        \cline{2-5}
        & \multicolumn{2}{c|}{All-search} & \multicolumn{2}{c}{Indoor-search}\\ 
        \cline{2-5}
    & R1 & mAP & R1 & mAP\\
        \hline
        \multicolumn{4}{l}{\textit{Supervised VI-ReID methods}} \\ \hline
        Zero-Pad \cite{SYSU-MM01} & 19.13 & 10.89 & 24.43 & 18.86 \\
        cmGAN \cite{cmGAN}  & 31.49 & 22.27 & 37.00 & 32.76\\
        AlignGAN \cite{AlignGAN}  & 51.50 & 33.90 & 57.10 & 45.30 \\
        cm-SSFT \cite{cm-SSFT}  & 63.40 & 62.00 & 73.00 & 72.40 \\
        MPANet \cite{MPANet}  & 75.58 & 62.91 & 84.22 & 75.11 \\
        FMCNet \cite{zhang2022fmcnet} & 73.44	& 56.06 & 78.86	& 63.82 \\
        CIFT \cite{CIFT} & 79.74 & \underline{75.56} & \underline{88.32} & \underline{86.42} \\
        PartMix \cite{PartMix} & \underline{80.54} & 69.84 & 87.99 & 79.95\\
        \hline
        \multicolumn{4}{l}
        {\textit{Unsupervised VI-ReID methods}} \\ \hline
        H2H \cite{H2H}& 30.31	& 19.12 & - & - \\
        DFC \cite{DFC}  & 44.12 & 28.36 & - & -\\
        MBCCM \cite{MBCCM} & 	\underline{57.73} & 39.78 & 62.87 & 52.80 \\
        CHCR \cite{CHCR} & 50.12 & \underline{42.17} & - & - \\
        \hline
    \textbf{Ours} & \textbf{71.35} & \textbf{52.18} & \textbf{76.99} & \textbf{64.03} \\
    \hline	
    \end{tabular}
    \end{adjustbox}
\label{tab:multi-shot}
\end{table}

\begin{table*}[!htbp]
    \centering	
    \caption{Ablation study on individual components of our method on SYSU-MM01 and RegDB.}
    \begin{adjustbox}{max width=\textwidth}
    \footnotesize
    \begin{tabular}{c|c|ccc|ccc|ccc|ccc}
    \hline
        \multicolumn{1}{c|}{
        \multirow{3}{*}{Index}} & 
        \multicolumn{1}{c|}{
        \multirow{3}{*}{Method}} &
    \multicolumn{6}{c|}{SYSU-MM01} & 
        \multicolumn{6}{c}{RegDB}\\
        \cline{3-14}
        & & \multicolumn{3}{c|}{All-search} & \multicolumn{3}{c|}{Indoor-search} &
        \multicolumn{3}{c|}{Visible-to-Infrared} & \multicolumn{3}{c}{Visible-to-Infrared}\\ 
        \cline{3-14}
    & & R1 & mAP & mINP & R1 & mAP & mINP & R1 & mAP & mINP &R1 & mAP & mINP\\
        \hline
        1 & Baseline & 41.76 & 38.57 & 24.32 & 45.83 & 54.44	& 50.00	& 50.58	& 48.78	& 35.62	& 51.17 & 47.24 & 33.80 \\
        \hline
        \multicolumn{12}{l}{\textit{The following methods are all based on the MULT module}} \\ 
        \hline
        2 & Baseline + $\mathcal{L}_{CM}$ & 60.02	&54.92	&39.18	&60.66	&66.92	&62.24	&83.74	&75.55	&58.26	&85.29	&75.08	&55.50\\
        3 & Baseline + $\mathcal{L}_{IM}$ + $\mathcal{L}_{CM}$ &60.18	&55.52	&39.86	&60.48	&67.39	&62.94	&85.87	&76.08	&58.05	&84.13	&74.36	&55.03 \\
        4 & Baseline + $\mathcal{L}_{OCLR}$ &54.34	&51.03	&36.21	&55.85	&62.68	&58.12
        & 85.87	&78.72	&65.41	&87.86	&79.48	&61.70 \\
        5 & Baseline + $\mathcal{L}_{IM}$ + $\mathcal{L}_{OCLR}$ &62.35	&58.22	&43.10	&64.77	&70.89	&66.38	&88.40	&81.20	&67.08	&88.93	&80.27	&62.73 \\
        6 & Baseline + $\mathcal{L}_{CM}$ + $\mathcal{L}_{OCLR}$ & 62.70 &58.25	&43.11	&64.06	&70.50	&66.11	&87.77	&80.07	&66.40	&88.16	&79.31	&62.68
        \\
        7 & Baseline + $\mathcal{L}_{IM}$ + $\mathcal{L}_{CM}$ + $\mathcal{L}_{OCLR}$ & \textbf{64.77} & \textbf{59.23} & \textbf{43.46} & \textbf{65.34} & \textbf{71.46} & \textbf{67.83} & \textbf{89.95} & \textbf{82.09} & \textbf{67.29} & \textbf{90.78} & \textbf{82.25} & \textbf{65.38} \\
    \hline	
    \end{tabular}
    \end{adjustbox}
\label{tab:ablation}
\end{table*}

\subsection{Comparison with State-of-the-Art Methods}

To demonstrate the efficiency of our proposed methods, we compare our method with state-of-the-art methods under three relevant settings on SYSU-MM01 and RegDB datasets, $i.e.$, supervised VI-ReID methods, unsupervised single-modality ReID methods, and unsupervised VI-ReID methods. The results are shown in Tab.\ref{tab:sota_comparison} and Tab.\ref{tab:multi-shot}.

\textbf{Comparison with Supervised VI-ReID Methods.} Our proposed method achieves competitive performance with supervised method VCD+VML \cite{VCD+VML} on SYSU-MM01, and outperform several supervised methods including Zero-Pad \cite{SYSU-MM01}, AlignGAN \cite{AlignGAN}, cm-SSFT \cite{AlignGAN}, DDAG \cite{DDAG} and AGW \cite{agw}.
Our method surpasses most supervised VI-ReID methods on the RegDB dataset and demonstrates excellent performance close to state-of-the-art methods.
The results reveal the significant developmental potential of unsupervised VI-ReID, yet there remains a considerable performance gap compared to the state-of-the-art supervised VI-ReID methods.

\textbf{Comparison with Conventional Unsupervised Single-Modality ReID Methods.} 
The results in Tab.\ref{tab:sota_comparison} indicate that unsupervised single-modality methods cannot effectively address the large modality-gap in cross-modality scenarios
Our method outperforms the state-of-the-art unsupervised single-modality ReID method Cluster-Contrast \cite{ClusterContrast} by a large margin of 37.23\% mAP and 44.61\% Rank-1 on SYSU-MM01, 
emphasizing the necessity of proposing cross-modality label association strategies for VI-ReID. 

\textbf{Comparison with Unsupervised VI-ReID Methods.} As reported in Tab.\ref{tab:sota_comparison}, our method outperforms the state-of-the-art GUR \cite{GRU} by 2.24\% mAP, 3.82\% Rank-1 on SYSU-MM01 (All-search) under the single-shot setting, and 11.86\% mAP, 16.04\% Rank-1 on RegDB (Visible-to-infrared). 
As shown in Tab.\ref{tab:multi-shot}, our method showcases remarkable performance, surpassing the state-of-the-art CHCR \cite{CHCR} by 10.01\% mAP and 21.23\% Rank-1 on SYSU-MM01 under the multi-shot setting (All-search). (Note that PGM \cite{PGM}, GRU \cite{GRU}, among others, do not provide results for the multi-shot setting on SYSU-MM01.)
The proposed method has three main advantages: 
(1) Unlike H2H \cite{H2H} and OTLA \cite{OTLA}, our method does not require additional datasets or camera labels for training.
(2) Our MULT associates cross-modality labels from a novel structural consistency perspective and can be applied to other unsupervised cross-domain tasks.
(3) Our OCLR is a plug-and-play module that can be employed in other unsupervised memory-based methods to handle label noise.

\subsection{Ablation Study}\label{ablation}

To validate the effectiveness of each component of our method, we conduct ablation experiments on SYSU-MM01 and RegDB datasets, as shown in Tab.\ref{tab:ablation}. 
Our baseline consists of a two-stream ResNet trained under the DCL framework \cite{ADCA}. 
All experiments are conducted with our MULT module. 
We do not perform an ablation study on $\mathcal{L}_{IM}$ as it does not address the problem of modality discrepancy.

\textbf{Effectiveness of MULT.} 
The MULT module delivers a +16.35\% mAP and +18.26\% Rank-1 improvement by directly incorporating $\mathcal{L}_{CM}$ (see $1^{st}$ row and $2^{nd}$ row in Tab.\ref{tab:ablation}) and achieves +20.66\% mAP and +23.01\% Rank-1 improvement when trained within our entire AMIRL framework on SYSU-MM01 (see $1^{st}$ row and $7^{th}$ row in Tab.\ref{tab:ablation}). 
To further evaluate the effectiveness of MULT, we compare MULT with other cross-modality label association methods in Tab.\ref{tab:compare with methods}, including PGM \cite{PGM}, MBCCM \cite{MBCCM}, CLU \cite{GRU} and DOTLA \cite{DOTLA}.
Specifically, we integrate the above methods into our AMIRL framework.
The results demonstrate our MULT provides higher-quality associations than other methods,  
achieving performance exceeding DOTLA \cite{DOTLA} by 4.19\% mAP without OCLR, and 3.88\% mAP by integrating our OCLR module. 
The core effectiveness of our MULT lies in the generated soft pseudo-labels containing richer contextual structural information within the feature space, which promotes the network's learning of shared fine-grained features both within clusters and across clusters.

\begin{table*}[!htbp]
    \centering	
    \caption{Comparison with other cross-modality label association methods for unsupervised VI-ReID on SYSU-MM01. All experiments are based on our AMIRL framework (Sec.\ref{MIRL}).}
    \begin{adjustbox}{max width=\textwidth}
    \footnotesize
    \begin{tabular}{c|c|ccc|ccc}
    \hline 
        \multicolumn{1}{c|}{
        \multirow{2}{*}{Method}} &
        \multicolumn{1}{c|}{
        \multirow{2}{*}{Venue}} &
    \multicolumn{3}{c|}{All-search} & 
        \multicolumn{3}{c}{Indoor-search}\\
        \cline{3-8}
    & & R1 & mAP & mINP & R1 & mAP & mINP \\
        \hline
        Baseline & - & 41.76 & 38.57 & 24.32 & 45.83 & 54.43 & 50.00 \\
        \hline
        \multicolumn{5}{l}{\textit{AMIRL (Sec.\ref{MIRL}) w/o OCLR (Sec.\ref{OCLR})}} \\ 
        \hline
        w/ PGM \cite{PGM} & CVPR-23 & 50.85 & 45.50 & 30.64 & 51.31 & 59.73 & 55.34 \\
        w/ BCCM \cite{MBCCM} & MM-23 & 51.35 & 46.61 & 31.97 & 53.44 & 61.28 & 56.66 \\
        w/ CLU \cite{GRU} & ICCV-23 & 57.09 & 48.96 & 31.88 & 57.11 & 64.51 & 60.32 \\
        w/ DOTLA \cite{DOTLA} & MM-23 & 57.17 & 51.33 & 36.07 & 55.16 & 62.53 & 57.87 \\
        \textbf{w/ MULT (Ours)} & -& \textbf{60.18} & \textbf{55.52} & \textbf{39.86} & \textbf{60.48} & \textbf{67.39} &
        \textbf{62.94}\\
        \hline
        \multicolumn{5}{l}{\textit{AMIRL (Sec.\ref{MIRL}) w/ OCLR (Sec.\ref{OCLR})}} \\ 
        \hline
        w/ PGM \cite{PGM} & CVPR-23 & 56.43 & 52.31 & 37.00 & 59.28 & 66.04 & 61.55 \\
        w/ BCCM \cite{MBCCM} &MM-23& 58.46 & 53.69 & 37.99 & 58.84 & 66.56 & 62.25\\
        w/ CLU \cite{GRU} &ICCV-23 & 61.40 & 54.18 & 38.11 & 62.91 & 68.32 & 64.08 \\
        w/ DOTLA \cite{DOTLA} &MM-23 &  61.53 & 55.35 & 39.96 & 60.82 & 67.60 & 63.06 \\
        \textbf{w/ MULT (Ours)} & -&\textbf{64.77} & \textbf{59.23} & \textbf{43.46} &
        \textbf{65.34} & 
        \textbf{71.46} & 
        \textbf{67.83}\\
    \hline	
    \end{tabular}
    \end{adjustbox}
    \vspace{-2mm}
\label{tab:compare with methods}
\end{table*}

\begin{table}[!htbp]
    \centering	
    \caption{Ablation study for the auxiliary memory.}
    \begin{adjustbox}{max width=0.48\textwidth}
    \footnotesize
    \begin{tabular}{c|ccc|cc}
    \hline 
        \multicolumn{1}{c|}{
        \multirow{2}{*}{Method}} &
    \multicolumn{3}{c|}{All-search} & 
        \multicolumn{2}{c}{Indoor-search}\\
        \cline{2-6}
    & R1 & mAP & mINP & R1 & mAP \\
        \hline
        w/ Intra& 63.77 & 58.61 & 43.04 & 64.54 & 71.38 \\
        w/ Aux& 63.06 & 58.20 & 42.65 & 65.06 & 71.07 \\
        Intra \& Aux & \textbf{64.77} & \textbf{59.23} & \textbf{43.46} & \textbf{65.34} & \textbf{71.46} \\
    \hline	
    \end{tabular}
    \end{adjustbox}
    \vspace{-4mm}
\label{tab:ablation for aux}
\end{table}

\begin{table}[!htbp]
    \centering	
    \caption{Ablation study for the alternative scheme.}
    \begin{adjustbox}{max width=0.48\textwidth}
    \footnotesize
    \begin{tabular}{c|ccc|cc}
    \hline 
        \multicolumn{1}{c|}{
        \multirow{2}{*}{Method}} &
    \multicolumn{3}{c|}{All-search} & 
        \multicolumn{2}{c}{Indoor-search}\\
        \cline{2-6}
    & R1 & mAP & mINP & R1 & mAP \\
        \hline
        w/ MIRL& 64.05 & 58.44 & 43.06 & 64.86 & 70.90 \\
        w/ AMIRL & \textbf{64.77} & \textbf{59.23} & \textbf{43.46} & \textbf{65.34} & \textbf{71.46} \\
    \hline	
    \end{tabular}
    \end{adjustbox}
    \vspace{-3mm}
\label{tab:ablation for alternative}
\end{table}

\textbf{Effectiveness of OCLR.} The OCLR module provides a performance gain of +12.46\% mAP and +12.58\% Rank-1 when directly adding it to our baseline (see $1^{st}$ row and $4^{th}$ row in Tab.\ref{tab:ablation}). When used in conjunction with our AMIRL framework, the OCLR further improves the performance of +3.71\% mAP and +4.59\% Rank-1 (see $3^{rd}$ row and $7^{th}$ row in Tab.\ref{tab:ablation}). 
The OCLR module significantly mitigates the impact of pseudo-label noise while narrowing the modality discrepancy.
OCLR utilizes ever-evolving prototypes for online refinement, avoiding overfitting static noisy labels.
We also integrate our OCLR into other methods (see upper and lower in Tab.\ref{tab:compare with methods}) based on our AMIRL framework. The results indicate that our OCLR improves performance when collaborating with other cross-modality association methods. 
{The proposed OCLR has the following advantages compared to the unsupervised single-modality ReID method MMT \cite{MMT}: (1) MMT focuses on consistency in predictions from different augmentations of the same instance, which does not consider the cross-modality interactions. Conversely, our OCLR emphasizes consistency in relationships between instances and multi-modality prototypes, 
which enhances cross-modality interactions and prompts learning modality-shared information.
Therefore, the proposed OCLR is more suitable for the USL-VI-ReID task.
(2) MMT refines predictions using the moving average model, while OCLR conducts refinement across multi-memories without storing past models, significantly reducing GPU memory consumption.}

\begin{figure*}
\centering
\includegraphics[width=1.0\textwidth]{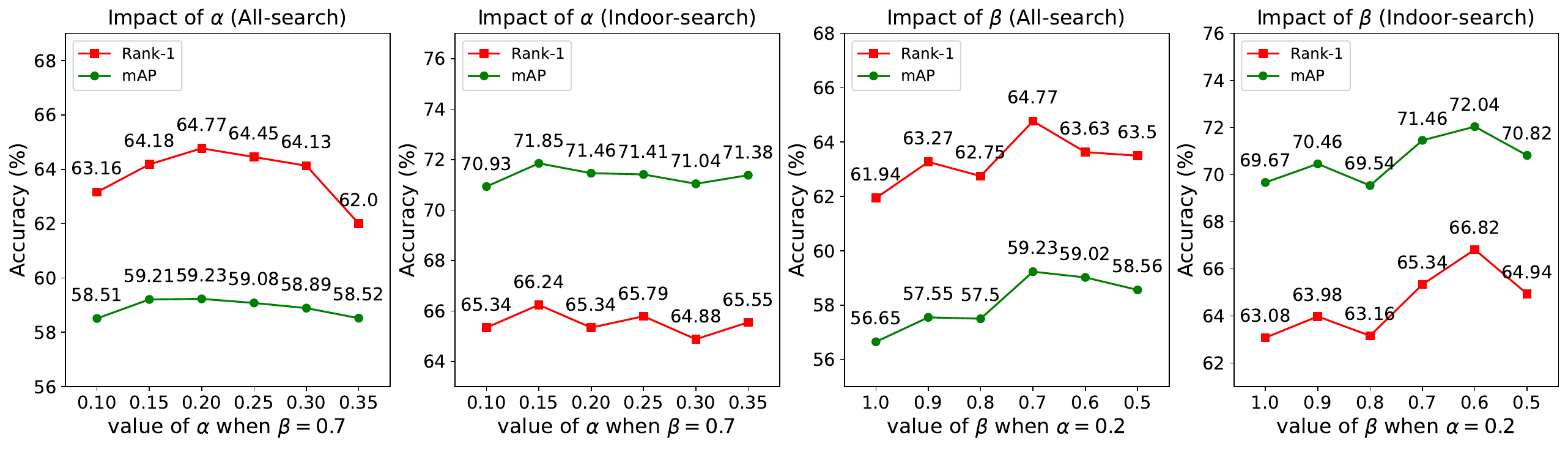}
\caption{Parameter analysis of $\alpha$ and $\beta$ on SYSU-MM01.}\label{parameter}
\vspace{-3mm}
\end{figure*}

\begin{table*}[!htbp]
    \centering	
    \caption{Comparison with various designs for modeling $\mathbf{S}^{he}$.}\label{Tab_fig:S_he}
    \footnotesize
    \begin{tabular}{c|ccc|ccc}
    \hline
        \multicolumn{1}{c|}{
        \multirow{2}{*}{Method}} & \multicolumn{3}{c|}{SYSU-MM01 (All-search)} & \multicolumn{3}{c}{SYSU-MM01 (Indoor-search)} \\ 
        \cline{2-7}
    & R1 & mAP & mINP & R1 & mAP & mINP\\
    \hline
    Jaccrad & 61.08 & 55.69 & 39.20	& 62.63	&68.49& 63.78\\
    Cosine (k=20) & 62.19 & 57.97 & 42.38 & 64.54&	70.88 & 66.57\\
    Cosine (k=30) & 61.73 & 57.66 & 41.91 & 64.65 & 70.59 & 66.08\\
    OT problem & \textbf{64.77} & \textbf{59.23} & \textbf{43.46} & \textbf{66.67} & \textbf{71.87} & \textbf{67.20}\\
    \hline	
    \end{tabular}
    \vspace{-4mm}
\label{tab: different dist}
\end{table*}

\textbf{Effectiveness of AMIRL.} Our AMIRL achieves +0.60\% mAP improvement without OCLR (see $2^{nd}$ row and $3^{rd}$ row in Tab.\ref{tab:ablation})
and achieves +2.41\% mAP with OCLR (see $6^{th}$ row and $7^{th}$ row in Tab.\ref{tab:ablation})
compared to leveraging $\mathcal{L}_{CM}$ alone. The results illustrate the importance of incorporating intra-modality contrastive loss $\mathcal{L}_{IM}$ into our framework,
especially when training with our OCLR module. 
Moreover, we conduct the ablation experiment for the auxiliary intra-modality memory and the alternative training scheme, as shown in Tab.\ref{tab:ablation for aux} and Tab.\ref{tab:ablation for alternative}. 
In Tab.\ref{tab:ablation for aux},
``w/ Intra" denotes only intra-modality memory banks $\tilde{\mathbf{M}}^v$ and $\tilde{\mathbf{M}}^r$ are involved in $\mathcal{L}_{IM}$ during training, 
and ``w/ Aux" 
denotes only auxiliary memory banks $\hat{\mathbf{M}}^r$ and $\hat{\mathbf{M}}^v$ are utilized. 
The results indicate that the cross-modality pseudo-labels also contribute to intra-modality learning. 
The improvement is attributed to these two types of pseudo-labels revealing two distinct intra-modality structures in feature space, thus correcting each other mutually. 
In Tab.\ref{tab:ablation for alternative}, ``MIRL" denotes the contrastive learning framework without the alternative scheme (AMIRL) for V-based training and R-based training.
The results demonstrate that the alternative scheme further enhances the performance.
This is achieved by alternating the utilization of labels from a single-directional MULT (V2R or R2V) to supervise model training across different epochs, thereby effectively mitigating the issue of pseudo-label inconsistency between V2R and R2V MULT. 
{Compared to a basic framework, MSMA \cite{MBCCM}, we construct an auxiliary memory bank to learn the structure of cross-modality pseudo-labels, this strategy fully exploits the pseudo-labels from MULT. Furthermore, we propose to alternate V-based training and R-based training. Such an alternative training scheme mitigates the side-effects of noisy labels from the inconsistent pseudo-labels output of the V2R / R2V MULT.
our AMIRL makes more effective utilization of the cross-modality pseudo-labels $\tilde{\mathbf{y}}^v$ and $\tilde{\mathbf{y}}^r$, thus is more suitable for our MULT label association algorithm.}



\subsection{Further Analysis}\label{exp_acc}

\textbf{Analysis for heterogeneous affinities modeling.} 
The Optimal Transport (OT) problem in Eq.\ref{OT_problem} is formulated to model heterogeneous affinities $\mathbf{S}^{he}$ from the perspective of the transition between instances from two different distributions.
The constraint in Eq.\ref{OT_problem} ensures that $N^v$ visible instances' affinities are uniformly distributed to $N^r$ infrared instances in the transport plan, and vice versa.
This design, compared to other alternative designs, prevents degenerated solutions where few infrared instances hold extremely high affinities with the majority of visible instances. 
We conducted a comparative analysis of various designs for modeling $\mathbf{S}^{he}$, as shown in Tab.\ref{Tab_fig:S_he}. 
``Jaccard'' denotes the cross-modality Jaccard Similarity-based affinities and ``Cosine (k=K)'' denotes the cosine-similarity-based knn-affinities, 
K denotes the number of knn-neighbors for constructing knn-graph.
The results confirmed the rationality of our choice.

\textbf{Parameter Analysis. }The proposed method includes two parameters in MULT, $i.e.$, $\alpha$ in Eq.\ref{Q_total_v} and $\beta$ in Eq.\ref{soft}. 
$\alpha$ is a trade-off parameter between the self-consistency terms and the heterogeneous consistency terms.
$\beta$ is the parameter to control the smoothness of the soft pseudo-labels from MULT.
To investigate the impact of these two parameters, we varied their values, as depicted in Figure \ref{parameter}. 
We find when $0.15 \leq \alpha \leq 0.3$ and $0.6 \leq \beta \leq 0.7$, the model achieves relatively outstanding performance. We set $\alpha=0.2$ and $\beta=0.7$ based on the experiments.

\begin{table*}[!htbp]
    \centering	
    \caption{Comparison with other methods on FMI/ARI.}
    \footnotesize
    \begin{tabular}
    {c|cccc|cccc}
    \hline 
    \multicolumn{1}{c|}{\multirow{2}{*}{Method}} &
    \multicolumn{4}{c|}{FMI metric (\%)} &
    \multicolumn{4}{c}{ARI metric (\%)}\\
    \cline{2-9}
     & RGB ($\tilde{y}^v$) & IR ($\tilde{y}^r$) & $\tilde{y}^v \& \hat{y}^r$& $\hat{y}^v \& \tilde{y}^r$ & RGB ($\tilde{y}^v$) & IR ($\tilde{y}^r$) & $\tilde{y}^v \& \hat{y}^r$& $\hat{y}^v \& \tilde{y}^r$ \\
    \cline{1-9} 
    PGM (CVPR23) &53.57	& 82.03	& 56.51&	52.05& 51.20 & 81.84 & 56.30 & 50.34 \\
    BCCM (MM23) & 51.99 & 83.33 & 60.20 & 52.61 & 50.03 & 83.20 & 52.29 & 44.92 \\
    OTLA (ECCV22) & 58.39 & 84.40 & 55.57 & 51.77 & 56.40 & 84.08 & 55.21 & 49.29 \\
    MULT (Ours) & \textbf{65.86} & \textbf{86.61} & \textbf{69.32} & \textbf{62.87} & \textbf{63.74} & \textbf{86.48} & \textbf{69.18} & \textbf{61.07} \\
    \hline
    \end{tabular}
\label{Tab_fig:ARI}
\end{table*}

\begin{table*}[!htbp]
    \centering	
    \caption{Ablation study on CA \cite{CA} in the proposed OCLR on SYSU-MM01.}
    \footnotesize
    \begin{tabular}{c|ccc|ccc}
    \hline
        \multicolumn{1}{c|}{
        \multirow{2}{*}{Method}} & \multicolumn{3}{c|}{SYSU-MM01 (All-search)} & \multicolumn{3}{c}{SYSU-MM01 (Indoor-search)} \\ 
        \cline{2-7}
    & R1 & mAP & mINP & R1 & mAP & mINP\\
    \hline
    Ours w/o OCLR & 60.18 & 55.52 & 39.18	& 60.66	&66.92& 62.24\\
    Ours w/ OCLR (w/o CA) & 61.98 & 57.44 & 42.05 & 63.01 & 69.62 & 65.42\\
    Ours w/ OCLR (w/ CA) &\textbf{64.77} & \textbf{59.23} & \textbf{43.46} &
    \textbf{65.34} & 
    \textbf{71.46} & 
    \textbf{67.83}\\
    \hline	
    \end{tabular}
    \vspace{-2mm}
\label{tab: CA ablation}
\end{table*}

\begin{table*}[!t]
    \centering	
    \caption{Ablation study on LTG \cite{LTG} on SYSU-MM01.}
    \footnotesize
    \begin{tabular}{c|ccc|ccc}
    \hline
        \multirow{2}{*}{Method} & \multicolumn{3}{c|}{RegDB (Visible-to-Infrared)}
        & \multicolumn{3}{c}{SYSU-MM01 (All-search)} \\ 
        \cline{2-7}
    & R1 & mAP & mINP & R1 & mAP & mINP\\
    \hline
    GUR$^\ast$(ICCV23,SOTA) & 73.91 & 70.23 & 58.88 & 60.95 & 56.99 & 41.85 \\
    Ours w/o LTG & 88.20 & 81.21 & 66.24 & 62.57 & 57.90 & 42.25  \\
    Ours w/ LTG & \textbf{89.95} & \textbf{82.09} & \textbf{67.29} & \textbf{64.77} & \textbf{59.23} & \textbf{43.46} \\
    \hline	
    \end{tabular}
\label{tab:ablation on LTG}
\end{table*}

\begin{figure}
\centering
\includegraphics[width=0.48\textwidth]{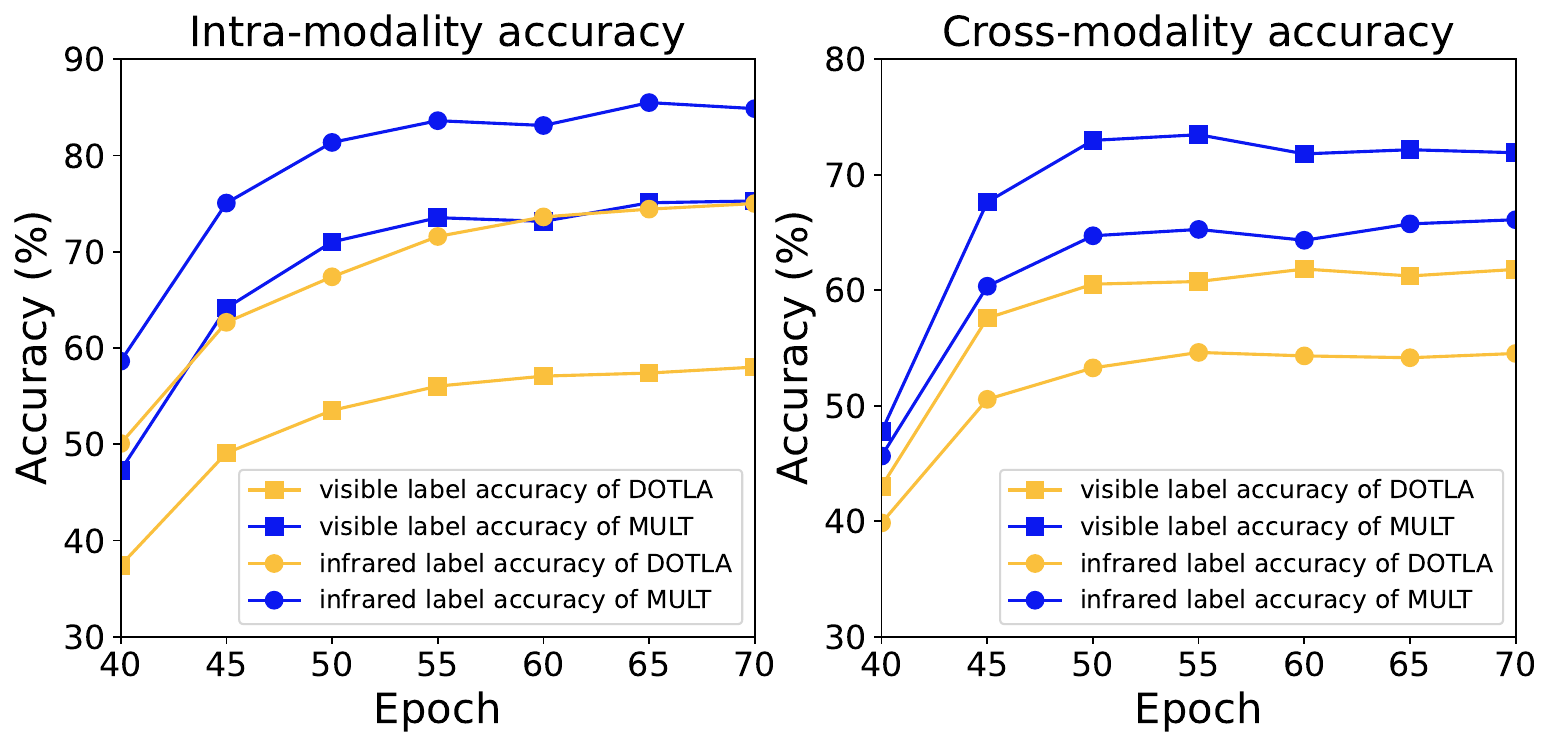}
\caption{Accuracy of intra-modality and cross-modality positive pairs found by pseudo-labels on SYSU-MM01.}\label{acc}
\vspace{-2mm}
\end{figure}

\begin{figure}
\centering
\includegraphics[width=0.48\textwidth]{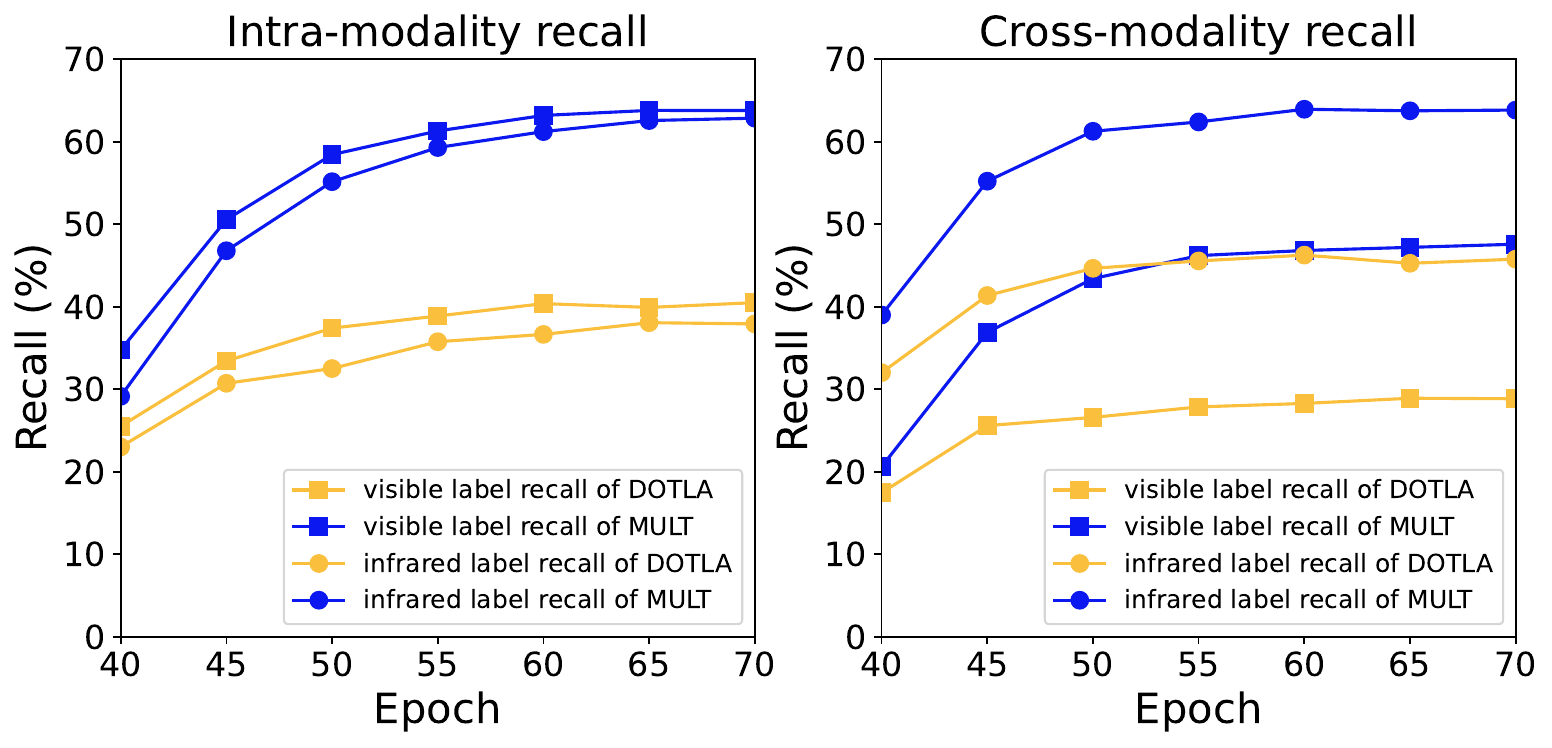}
\caption{Recall of intra-modality and cross-modality positive pairs found by pseudo-labels on SYSU-MM01.}\label{recall}
\vspace{-3mm}
\end{figure}

\textbf{Pseudo-label Accuracy Analysis.} To quantify the quality of the pseudo-labels, we calculate the accuracy/recall of positive instance pairs found by the pseudo-labels during training, as shown in Fig.\ref{acc} and Fig.\ref{recall}. 
We denote the ground-truth hard labels as $y^e(gt)$, where $e = \{v, r\}$ indicates the visible and infrared modality, respectively. 
We utilize the hard form $y^e = \{\tilde{y}^v, \tilde{y}^r, \hat{y}^v, \hat{y}^r\}$ of the soft pseudo-labels from MULT when quantifying the quality.
In Fig.\ref{acc}, `Intra-modality visible label accuracy' ($IntraAcc^v$) indicates the accuracy of the visible intra-modality positive pairs found by the cross-modality visible labels $\hat{{y}}^v$. This can be achieved by computing 
the proportion of the ground-truth positive pairs within all instance pairs that share the same pseudo-label:

\vspace{-4mm}
\begin{equation}\label{intra_acc}
    IntraAcc^v = \frac{\sum_{i=1}^{N^v} \sum_{j=1}^{N^v} \mathbf{1} (\hat{{y}}^v_i = \hat{{y}}^v_j) \mathbf{1} ({y}^v_i(gt) = {y}^v_j(gt))}
    {\sum_{i=1}^{N^v} \sum_{j=1}^{N^v} \mathbf{1} (\hat{{y}}^v_i = \hat{{y}}^v_j)}.
\end{equation}
\vspace{-4mm}

\noindent Similarly, `Cross-modality visible label accuracy' ($CrossAcc^v$) indicates the accuracy of the cross-modality positive pairs found by $\tilde{\mathbf{y}}^v$ and $\hat{\mathbf{y}}^r$ from the V2R MULT. It can be similarly formulated as follows:

\vspace{-3mm}
\begin{equation}\label{cross_acc}
    CrossAcc^v = \frac{\sum_{i=1}^{N^v} \sum_{j=1}^{N^r} \mathbf{1} (\tilde{{y}}^v_i = \hat{{y}}^r_j) \mathbf{1} ({y}^v_i(gt) = {y}^r_j(gt))}
    {\sum_{i=1}^{N^v} \sum_{j=1}^{N^r} \mathbf{1} (\tilde{{y}}^v_i = \hat{{y}}^r_j)}.
\end{equation}
\vspace{-5mm}

\noindent The recall in Fig.\ref{recall} can be obtained similarly by computing the proportion of pairs that share the same pseudo-label within all ground-truth positive pairs.
We compare our MULT with DOTLA \cite{DOTLA}, and the results demonstrate that our MULT significantly 
facilitates the quality of pseudo labels. Simultaneously, the results indicate that as MULT and network training alternate, the network can continuously correct erroneous pseudo-labels.

To further measure the reliability of pseudo-labels derived from our MULT, we employ 
Fowlkes-Mallows Index (FMI) and Adjusted Rand Index (ARI) metrics to evaluate the pseudo-labels across various association methods, as shown in Tab.\ref{Tab_fig:ARI}. The FMI score indicates the geometric mean of the accuracy and recall, while the ARI score reflects the degree of overlap between the ground-truth label space and the pseudo-label space. The results illustrate the superiority of our MULT.

\begin{figure}
\centering
\includegraphics[width=0.48\textwidth]{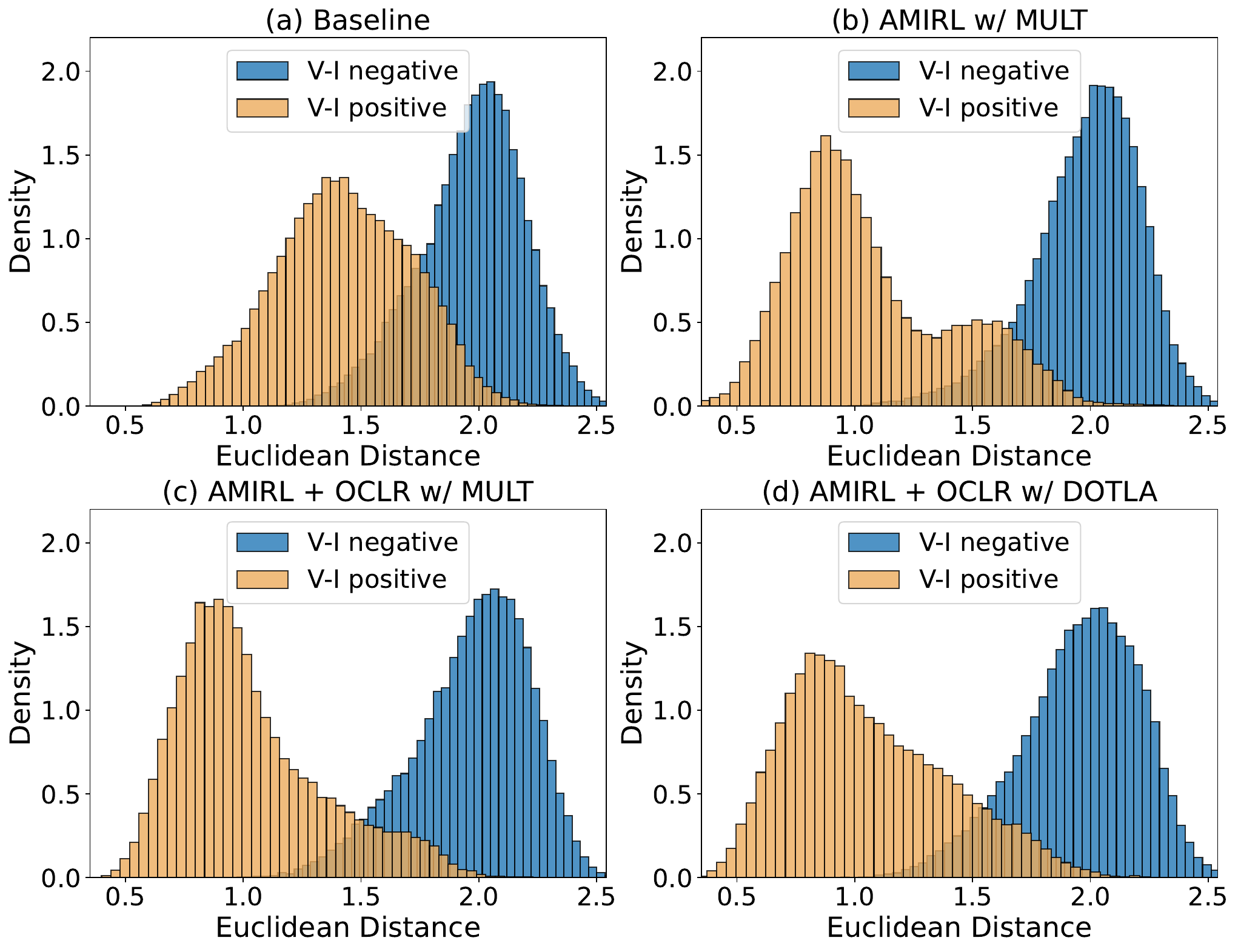}
\caption{The visualization of Euclidean distance distribution of randomly selected cross-modality positive and negative pairs.}\label{plot-dist}
\vspace{-5mm}
\end{figure}

\begin{figure*}[!t]
\centering
\includegraphics[width=1.01\textwidth]{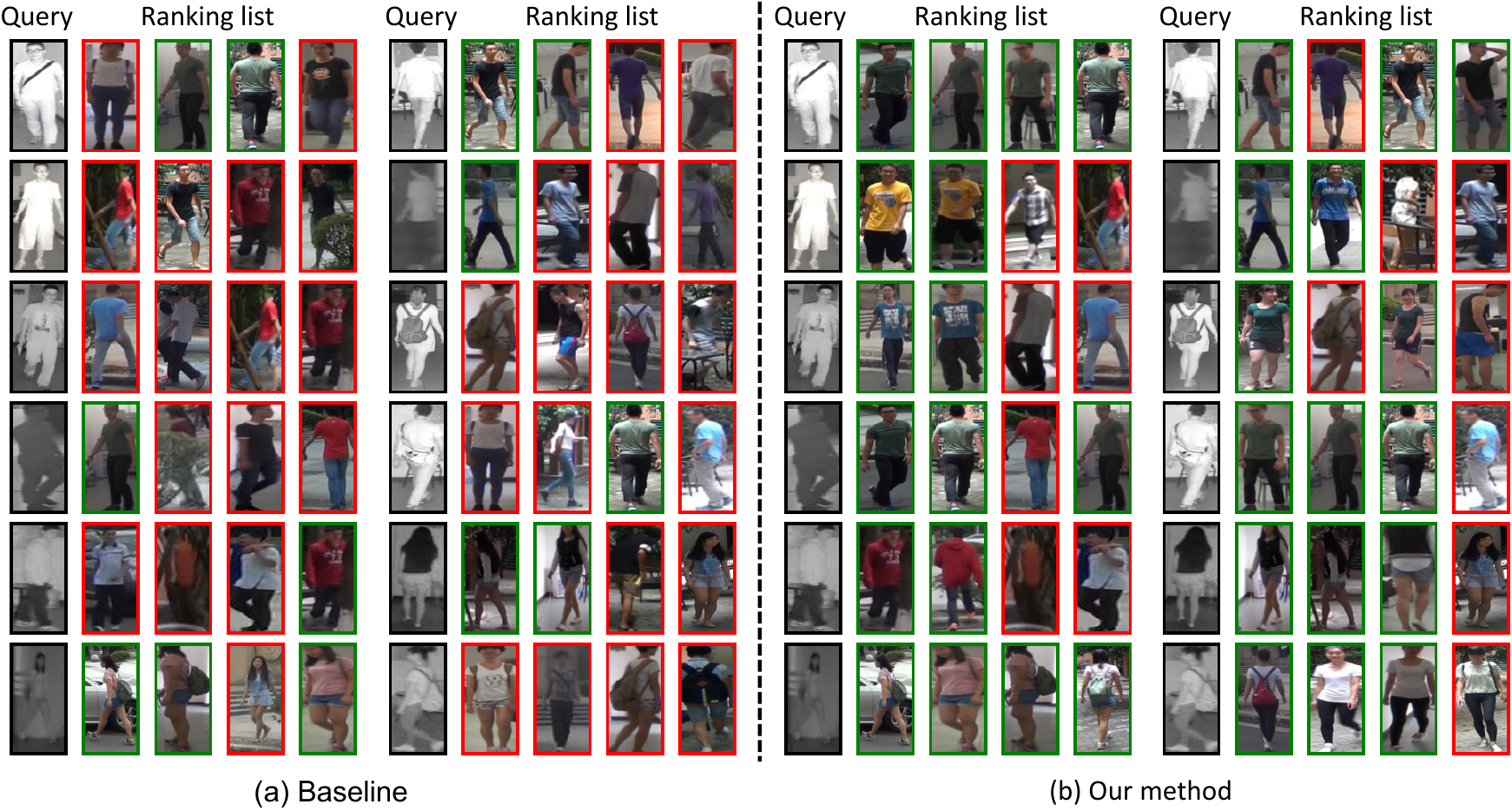}
\caption{Visualization of the ranking lists on the SYSU-MM01 dataset. The persons who are different from the query
persons are marked with red boxes, while those who are the same as the query are marked with green boxes. 
}\label{ranking-list}
\vspace{-3mm}
\end{figure*}

\textbf{Distribution Visualization.} We visualize the Euclidean distance distribution of randomly selected 50000 positive and negative visible-infrared pairs, as shown in Fig.\ref{plot-dist}. By sequentially integrating the modules ($i.e.$, MULT, AMIRL, and OCLR)
into the training framework, we observe a convergence of cross-modality positive pairs and a divergence of negative pairs, 
demonstrating the effectiveness of each component in our framework to address the modality discrepancy. 
We further visualize the distribution by integrating DOTLA into our framework in Fig.\ref{plot-dist} (d),
the results illustrate the superiority of MULT in comparison with DOTLA \cite{DOTLA}.

\begin{figure}[t]
\centering
\includegraphics[width=0.48\textwidth]{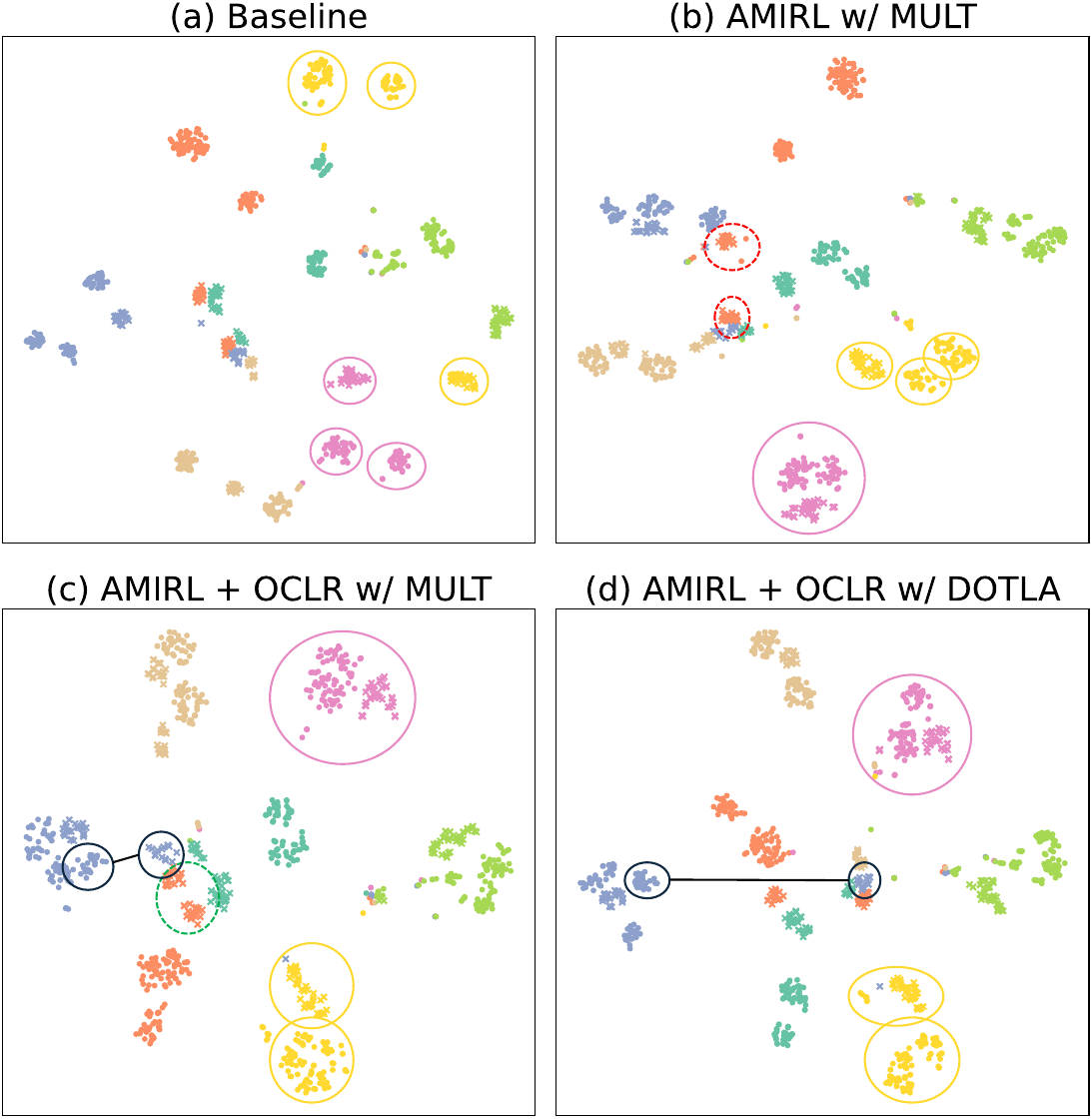}
\caption{The t-SNE \cite{t-sne} visualization of the feature of 7 randomly selected identities. Different colors mean different identities. Circle means the visible modality and cross means the infrared modality.}\label{t-sne}
\vspace{-3mm}
\end{figure}

\textbf{T-SNE Visualization.} We visualize the t-SNE \cite{t-sne} map of 7 randomly selected identities, as shown in Fig.\ref{t-sne}. 
In Fig.\ref{t-sne} (a) and Fig.\ref{t-sne} (b), it is evident that, owing to the precise label associations from MULT, cross-modality positive pairs exhibit increased proximity within the embedding space.
Fig.\ref{t-sne} (c) and Fig.\ref{t-sne}(d) provide a comparative analysis between our MULT and DOTLA \cite{DOTLA}. The results reveal the issue of homogeneous inconsistency in DOTLA, where images of the same identities are distributed into distinct clusters within the embedding space (black circles in Fig.\ref{t-sne} (d)). Our MULT effectively addresses this problem (black circles in Fig.\ref{t-sne} (c)) by generating pseudo-labels with low homogeneous consistency. 
The effectiveness of our OCLR is demonstrated when comparing Fig.\ref{t-sne} (b) and Fig.\ref{t-sne} (c).
Without OCLR, the network struggles to learn compact and discriminative representations due to the inevitable noisy labels (red dotted circles in Fig.\ref{t-sne} (b)). With OCLR, it shows more discriminative features that are well-distributed around the edge of the clusters (green dotted circles in Fig.\ref{t-sne} (c)). The results indicate that our OCLR module is robust to noise while facilitating modality alignment.

{\textbf{Analysis of the data augmentation strategies in the proposed method.} Data augmentations are widely-used in VI-ReID, we conduct ablation studies for the data augmentations included in our framework, $i.e.$, CA \cite{CA} and LTG \cite{LTG}. CA \cite{CA} is a commonly-used technology in existing USL-VI-ReID baselines \cite{MBCCM, PGM, ADCA, DOTLA, GRU}. We conduct ablation studies of the CA augmentation utilized in OCLR, as shown in Tab.\ref{tab: CA ablation}. As shown in Tab.\ref{tab: CA ablation}, the OCLR module is still effective when removing CA, which outperforms the framework without OCLR by 1.92\% mAP and 1.8\% Rank1. When incorporating the CA-augmented images into our OCLR, the performance further improves by 1.79\% mAP and 2.79\% Rank1. It can be attributed to the integration of CA bridges different modalities and promotes the model to better learn modality-shared features. 
We further study the improvement brought by the LTG augmentation, as shown in Tab.\ref{tab:ablation on LTG}.
The results show a slight drop in performance after removing LTG, but it still outperforms the state-of-the-art GUR$^\ast$.
The results also show that LTG further improves the model performance.
}

\textbf{Visualization of Ranking List.}
We visualize some of the ranking lists on the SYSU-MM01 dataset, as shown in Fig.\ref{ranking-list}. We compare our method with the baseline pretrained under the DCL framework \cite{ADCA}. The results demonstrate the effectiveness of our method in generating high-quality cross-modality supervision signals for training.

\section{Conclusion}

In this paper, we propose the Modality-Unified Label Transfer module to establish high-quality cross-modality pseudo-label associations for training from a novel structural consistency perspective. 
Our MULT lays emphasis on preserving both homogeneous and heterogeneous structure information in pseudo-label space by instance-wise affinities-guided label transfer.
To fully exploit the soft pseudo-labels from MULT, an Alternative Modality-Invariant Representation Learning framework is proposed based on both intra-modality and cross-modality contrastive learning. 
Furthermore, we introduce a straightforward yet effective plug-and-play Online Cross-memory Label Refinement module, which simultaneously mitigates the negative effects of noisy labels and facilitates modality alignment. 
Extensive experiments have demonstrated that our framework outperforms prior state-of-the-art methods. 
In the future, we will investigate more robust cross-modality label association methods based on our proposed MULT, which is the core issue of the unsupervised VI-ReID task.

\noindent\textbf{Data Availability Statement} 

\noindent\textbf{SYSU-MM01:} A signed dataset release
\href{https://github.com/wuancong/SYSU-MM01}{\textcolor{blue}{agreement}} must be send to \href{mailto:wuancong@gmail.com}{\textcolor{blue}{wuancong@gmail.com}} or \href{mailto:wuanc@mail.sysu.edu.cn}{\textcolor{blue}{wuanc@mail.sysu.edu.cn}} to obtain a download link.

\noindent\textbf{RegDB:} The dataset can be downloaded by submitting a copyright form to \href{http://dm.dongguk.edu/link.html}{\textcolor{blue}{http://dm.dongguk.edu/link.html}}.


%
%


\bibliographystyle{spbasic}
\bibliography{sn-bibliography}    

%
%

\end{sloppypar}
\end{document}